\documentclass{article}

\PassOptionsToPackage{numbers, compress}{natbib}

 \usepackage[preprint]{neurips_2026}

\usepackage[utf8]{inputenc} %
\usepackage[T1]{fontenc}    %
\usepackage{hyperref}       %
\usepackage{url}            %
\usepackage{booktabs}       %
\usepackage{amsfonts}       %
\usepackage{nicefrac}       %
\usepackage{microtype}      %
\usepackage{xcolor}         %
\usepackage{graphicx}
\usepackage{multirow}
\usepackage{makecell}
\usepackage{caption}
\usepackage{subcaption}
\usepackage{comment}
\usepackage{amsmath}    %
\usepackage{amsfonts}   %
\usepackage{amssymb}    %
\usepackage{bm}
\usepackage{algorithm}
\usepackage{algpseudocode}
\usepackage{adjustbox}
\usepackage{booktabs}
\usepackage{tikz}
\usepackage{pgf-pie}
\usetikzlibrary{positioning}

\usepackage{booktabs}
\usepackage{tabularx}
\usepackage{multicol}
\usepackage{multirow}
\usepackage{enumitem}
\usepackage{pgfplots}
\usepackage{placeins}
\usepackage[most]{tcolorbox}
\usepackage[table]{xcolor}

\title{PrimitiveVLA: Learning Reusable Motion Primitives for Efficient and Generalizable Robotic Manipulation}

\author{%
  \textbf{Yutai Li}$^{1,2,3,4}$,
  \textbf{Shaohui Peng}$^{5}$,
  \textbf{Jiaming Guo}$^{1,3}$,
  \textbf{Di Huang}$^{1,3}$,
  \textbf{Zihao Zhang}$^{2}$,
  \textbf{Yuxuan Guo}$^{1,4,6}$ \\
  \textbf{Yunkai Gao}$^{2}$,
  \textbf{Siming Lan}$^{2}$,
  \textbf{Ling Li}$^{5}$,
  \textbf{Xing Hu}$^{1,3}$,
  \textbf{Yunji Chen}$^{2,1,3,}$\thanks{Corresponding authors.} \\
  $^{1}$ State Key Lab of Processors, Institute of Computing Technology, CAS \\
  $^{2}$ Jiangsu Key Laboratory of AI for Industries, Institute of AI for Industries, CAS \\
  $^{3}$ University of Chinese Academy of Sciences \quad
  $^{4}$ Cambricon Technologies \\
  $^{5}$ Intelligent Software Research Center, Institute of Software, CAS \\
  $^{6}$ University of Science and Technology of China \\
  \texttt{\{liyutai23s, cyj\}@ict.ac.cn} \\
}

\begin{document}
\maketitle

\begin{abstract}
  Vision-Language-Action (VLA) models offer a promising paradigm for generalist robotic policies, yet their adaptation is hindered by \textbf{data inefficiency} and \textbf{poor generalization}. We argue that these bottlenecks stem from the prevailing \textbf{Direct Instruction-to-Control Mapping}, which forces models to memorize monolithic trajectories rather than reusable motion patterns, i.e., primitives. We propose \textbf{PrimitiveVLA}, a framework that shifts this paradigm toward a \textbf{Primitive-Centric Disassemble \& Assemble} paradigm. Supported by a shared \textbf{Multimodal Canonical Representation} (MCR), PrimitiveVLA unifies two phases: (1) \textbf{Fine-tuning-phase Disassembly}, which uses an automated pipeline to disassemble demonstrations into reusable primitives; and (2) \textbf{Inference-phase Assembly}, which employs a VLM-based planner and an LLM-generated switch module for robust closed-loop execution. By disassembling tasks into \textbf{reusable primitives}, PrimitiveVLA enables VLA models to learn invariant motion patterns instead of task-specific trajectories. Extensive experiments show that our framework improves data efficiency and achieves superior zero-shot generalization across unseen and long-horizon tasks.
\end{abstract}

\section{Introduction}
\label{sec:Introduction}

The convergence of Large Language Models (LLMs) and robotic control has spurred the rise of Vision-Language-Action (VLA) models. By projecting visual observations, instructions, and actions into a unified latent space, models like RT-2~\cite{brohan2023rt2} and OpenVLA~\cite{kim2024openvla} demonstrate impressive capabilities in open-world manipulation. Trained on internet-scale datasets (e.g., Open X-Embodiment~\cite{padalkar2023openx} and DROID~\cite{droid2024}), these generalist policies promise a ``foundation model'' moment for embodied AI.

\begin{figure*}[t] %
    \centering
    \includegraphics[width=\textwidth]{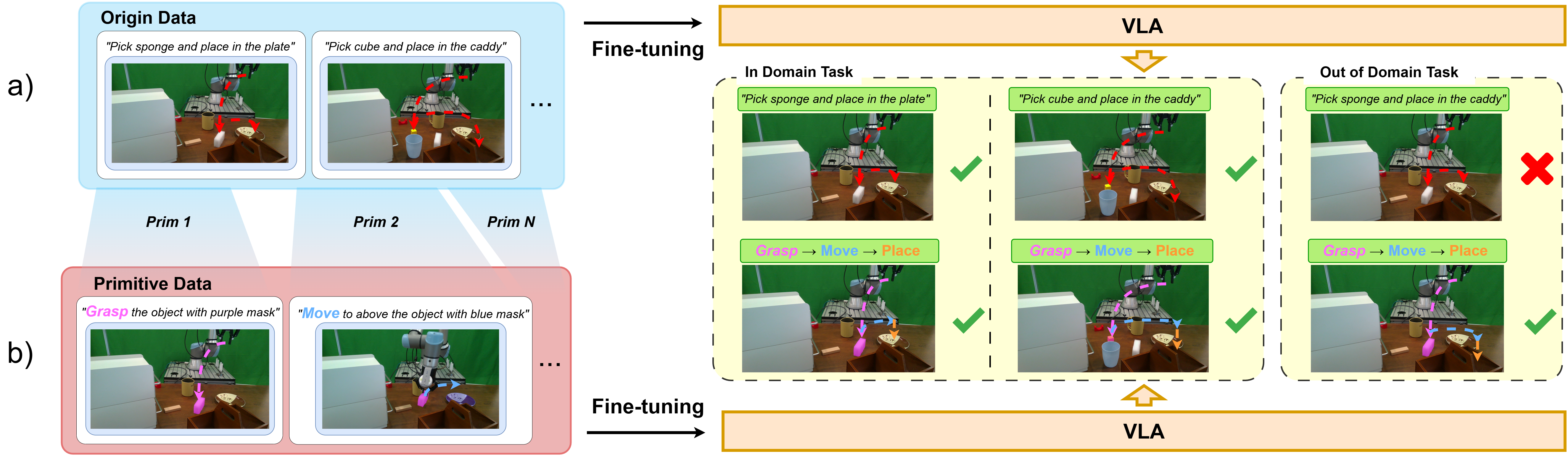}
   \caption{\textbf{Direct Instruction-to-Control Mapping vs. Primitive-Centric Disassemble \& Assemble.}
    (a) Conventional paradigms treat multi-stage tasks as indivisible units, leading to low data efficiency and limited generalization.
    (b) \textbf{PrimitiveVLA} \textbf{disassembles} multi-stage interactions into a library of reusable primitives. By mastering these reusable motion patterns, the model can flexibly \textbf{assemble} learned primitives to solve novel, unseen tasks via compositional generalization.}
    \label{fig:teaser}
    \vspace{-10pt}
\end{figure*}

However, adapting pre-trained VLAs to downstream domains via fine-tuning remains challenging. Despite vast pre-training, VLAs often exhibit brittle performance with \textbf{weak cross-task generalization}. Benchmarks like AGNOSTOS~\cite{zhou2025agnostos} reveal a significant struggle with novel object-skill combinations. We argue that this limitation is not only due to data volume but also a fundamental flaw in the prevailing \textbf{Direct Instruction-to-Control Mapping} paradigm in fine-tuning. As illustrated in Figure~\ref{fig:teaser}(a), this paradigm entangles distinct physical motion patterns within task-specific trajectories. By mapping high-level instructions directly to low-level control, the model \textbf{lacks structural constraints to decouple motion from context}, leading it to shortcut by memorizing scene-specific visual patterns instead of reusable physical motion patterns. Although recent $\pi_{0.7}$~\cite{intelligence2026pi07steerablegeneralistrobotic} decomposes tasks at the inference stage, its fine-tuning process maintains a tight coupling between semantic instructions and low-level action, adhering to the same Direct Instruction-to-Control Mapping paradigm. This makes data reuse difficult: knowledge from "opening a cabinet" is hard to transfer to "opening a microwave," leading to \textbf{low data efficiency} where every new task requires expensive new demonstrations.

To address these challenges, the \textbf{Primitive-Centric Disassemble \& Assemble} paradigm (Figure~\ref{fig:teaser}(b)) is proposed for VLA fine-tuning. This paradigm \textbf{disassembles} task trajectories into a finite set of \textbf{primitives}—shared, task-agnostic motion patterns such as \textit{Grasp} and \textit{Pull}. By treating these segmented primitives as independent training samples and re-representing their inputs, we decouple physical motion from specific task contexts to fine-tune the VLA. This re-regulation of data forces the VLA to shift its learning focus from memorizing task-specific scene correlations to mastering underlying motion patterns that are reusable across diverse tasks; for instance, "pull" a cabinet or a microwave now trains the same primitive. During execution, the VLA outputs continuous actions guided by a planner's primitive sequence, \textbf{assembling} these primitives to solve novel or long-horizon tasks, thereby significantly enhancing data efficiency and generalization capability.

However, realizing this paradigm requires overcoming several challenges:
\begin{enumerate}
\item \textbf{Disassembling during Fine-tuning (Label Scarcity):} Disassembling trajectories into primitives during fine-tuning is hindered by the fact that public datasets only provide task-level instructions. Manually annotating fine-grained primitive labels is expensive.
\item \textbf{Assembling during Inference (Control Complexity):} Effectively assembling discrete primitives into a coherent task during inference requires a robust mechanism to manage the execution flow and precisely trigger switching logic in closed-loop environments.
\item \textbf{Maintaining Invariance across Tasks (Contextual Interference):} To make the "disassemble-assemble" paradigm viable, primitives must remain consistent and reusable. However, task-specific details often interfere with primitive learning, whereas over-unification can remove task-relevant information, making primitives unreusable.
\end{enumerate}

Building on these, we propose \textbf{PrimitiveVLA}, which shifts VLA learning to the Disassemble \& Assemble paradigm. Our framework addresses these challenges through two phases supported by a shared Multimodal Canonical Representation (MCR). Specifically, to address label scarcity in \textbf{fine-tuning} (Sec.~\ref{sec:segmentation}), we develop an automated segmentation pipeline using VLMs and LLMs to disassemble trajectories into primitives without manual effort. To handle Control Complexity during \textbf{inference} (Sec.~\ref{sec:inference}), we design an assembly mechanism: a VLM planner sequences the primitives, while LLM-generated code manages the switching logic to assemble them into a coherent task. Finally, the \textbf{MCR} (Sec.~\ref{sec:representation}) resolves Contextual Interference by unifying instructions to align motion patterns and using object-centric masks to preserve task context. This ensures that primitives disassembled during fine-tuning remain consistent and adaptable for assembly during inference, enabling reliable cross-task reuse within this paradigm.

Our contributions are summarized as follows:

(1) \textbf{Primitive-Centric Disassemble \& Assemble Paradigm:} We introduce the Primitive-Centric Disassemble \& Assemble paradigm, shifting VLA learning from monolithic trajectories to reusable physical primitives. This allows models to disassemble tasks during fine-tuning and assemble them for inference to solve diverse problems.(2) \textbf{PrimitiveVLA Framework:} We propose PrimitiveVLA, a model-agnostic framework that unifies fine-tuning and inference via a shared Multimodal Canonical Representation (MCR). It leverages an automated segmentation pipeline for primitive disassembly and a hierarchical VLM-LLM mechanism for execution assembly.(3) \textbf{Experimental Evaluation:} Extensive experiments demonstrate that PrimitiveVLA significantly improves \textbf{data efficiency and cross-task generalization}. On Libero-90, our method boosts OpenVLA performance by \textbf{9.2\%} and achieves comparable results while \textbf{doubling the data efficiency}. Crucially, it achieves a \textbf{6$\times$ improvement} on unseen tasks for OpenVLA and elevates the SOTA ($\pi_{0.5}$) success rates on long-horizon tasks from 30.50\% to \textbf{80.25\%}.

\section{Related Work}
\subsection{Vision-Language-Action Models}
\looseness=-1 Vision-Language-Action (VLA) models have revolutionized generalist robot learning. Beyond the pioneering RT series~\cite{brohan2022rt1,brohan2023rt2} and VIMA~\cite{jiang2023vima}, open-weights models like OpenVLA~\cite{kim2024openvla}, Octo~\cite{octo_2023}, and RoboFlamingo~\cite{roboflamingo} have democratized access.
In parallel, researchers have explored diverse methodological enhancements: Diffusion Policy~\cite{chi2023diffusion} and RDT~\cite{rdt2024} investigate diffusion processes to improve action expressiveness, while VideoVLA~\cite{shen2025videovlavideogeneratorsgeneralizable} and CoT-VLA~\cite{Zhao_2025_CVPR} leverage generative visual embeddings to aid decision-making. 
Simultaneously, training paradigms have diversified, ranging from the optimized fine-tuning (OFT) recipe in OpenVLA-OFT~\cite{kim2025optimizing} to flow matching in the recent $\pi_0$ series~\cite{black2024pi0,intelligence2025pi05, intelligence2026pi07steerablegeneralistrobotic}.

However, regardless of architecture or training strategy, existing approaches predominantly follow a \textbf{Direct Instruction-to-Control Mapping} paradigm, training models on trajectories as indivisible units. We diverge from this via the Primitive-Centric Disassemble \& Assemble paradigm, which segments training data into \textbf{reusable primitives}. By decoupling physical motions from high-level task semantics during training, this approach unlocks superior data efficiency and generalization.

\subsection{Primitives and Hierarchical Robot Learning}
Hierarchical control often decomposes tasks into modular skills. While classical methods relied on options~\cite{sutton1999options} or DMPs~\cite{ijspeert2013dmp}, modern approaches focus on unsupervised skill discovery. 
Continuous latent models (SPiRL~\cite{pertsch2020spirl}, OPAL~\cite{ajay2020opal}) and vector quantization methods (BeT~\cite{shafiullah2022bet}, VQ-BeT~\cite{lee2024vqbet}, UAE~\cite{uae2024}) compress trajectories into compact codes.
However, lacking explicit semantic supervision, these statistical representations (e.g., latent $\bm{z}$) fail to capture the interpretable, reusable physical definitions required for VLA fine-tuning.

Alternatively, LLM-based planners like SayCan~\cite{ahn2022saycan} and Code as Policies~\cite{liang2023code} leverage semantics but operate only as high-level planners.
Recent decomposition attempts also face limitations: Pivot-R~\cite{pivotr} and Manipulate Anything~\cite{manipulateanything} predict segments using only instructions and images, failing to generate the precise, action-aligned segmentation needed for fine-tuning; while StARe VLA~\cite{starevla} utilizes decomposition primarily for RL rewards; Most recently, $\pi_{0.7}$~\cite{intelligence2026pi07steerablegeneralistrobotic} incorporates world models to introduce rich predictive information for hierarchical control, yet lacks explicit primitive-level decoupling.
In contrast, PrimitiveVLA extracts precise semantic primitives to directly fine-tune pre-trained VLAs, bridging high-level reasoning with capable low-level control.

\section{Preliminaries}

\subsection{Problem Formulation}
\label{sec:problem_formulation}

\noindent \textbf{Base Formulation.} 
We formulate robotic manipulation as a sequential decision-making problem. At step $t$, the policy observes a state tuple $(o_t, s_t)$, where $o_t=\{I^{(global)}_t, I^{(wrist)}_t\}$ are RGB images and $s_t \in \mathbb{R}^{7}$ is the proprioceptive state. The policy predicts actions $a_t \in \mathbb{R}^{7}$ (6-DoF delta pose and gripper action). A complete task demonstration as a trajectory $\tau = \{(o_t, s_t, a_t)\}_{t=1}^T$.

\noindent \textbf{Direct Instruction-to-Control Mapping vs. Primitive-Centric Disassemble \& Assemble.} 
Standard fine-tuning follows a Direct Instruction-to-Control Mapping, conditioning the policy on a high-level task instruction $l$ to optimize:
\begin{equation}
    \mathcal{L}_{task} = - \sum_{t=1}^{T} \log \pi_\theta(a_t \mid o_t, s_t, l).
    \label{eq:standard_loss}
\end{equation}
This entangles task semantics with low-level control, leading to scene-specific memorization. In contrast, our paradigm \textbf{disassembles} $\tau$ into primitive-aligned samples: $\tau \rightarrow \{(\tilde{o}_i, s_i, a_i, c_i)\}_{i=1}^N$, where $c_i$ is a \textbf{canonical primitive instruction} and $\tilde{o}_i$ is a \textbf{masked observation}. The VLA is trained to master these reusable patterns via:
\begin{equation}
    \mathcal{L}_{prim} = - \sum_{i=1}^{N} \log \pi_\theta(a_i \mid \tilde{o}_i, s_i, c_i).
    \label{eq:primitive_vla}
\end{equation}
This formulation decouples invariant physical motions from task contexts, transforming the VLA into a reusable primitives master.

\begin{table}[t]
    \centering
    \footnotesize
    \renewcommand{\arraystretch}{0.95} %
    \setlength{\tabcolsep}{2.5pt} %
    
    \vspace{-5pt}
    \caption{\textbf{The set of reusable primitives.}
    We define 11 primitives as reusable units for interactions.}
    \label{tab:primitive_library}
    \vspace{1em}
    \setlength{\tabcolsep}{4pt}
    \begin{tabular}{@{}ll p{0.5\linewidth}@{}}
        \toprule
        \textbf{Category} & \textbf{Primitive} & \textbf{Kinematic Definition} \\
        \midrule
        \multirow{4}{*}{\makecell[c]{\textit{Spatial} \\ \textit{Transport}}}
        & \textbf{Grasp} & Approaches, seizes target, and performs preliminary lift. \\
        & \textbf{Place} & Descends, releases object, and retreats vertically. \\
        & \textbf{Lift}  & Vertical translation ($+z$) while holding an object. \\
        & \textbf{Move}  & Planar translation ($xy$) maintaining object grasp. \\
        \midrule
        \multirow{3}{*}{\makecell[c]{\textit{Contact \&} \\ \textit{Interaction}}}
        & \textbf{Push}  & Slides object on surface to target without grasping. \\
        & \textbf{Pull}  & Drags or retracts object towards target position. \\
        & \textbf{Insert}& Aligns and inserts into a constrained slot. \\
        & \textbf{Press} & Applies downward force on an object or surface. \\
        \midrule
        \multirow{3}{*}{\makecell[c]{\textit{Orientation}}}
        & \textbf{Twist} & Rotates gripper (roll) for rotary mechanisms. \\
        & \textbf{Tilt}  & Reorients effector (pitch/yaw) to alter pose. \\
        & \textbf{Rotate}& Follows articulation trajectory (e.g., lid). \\
        \bottomrule
    \end{tabular}
    \vspace{-10pt} 
\end{table}

\subsection{Composable Primitives}
\label{sec:primitive_design}

To structure physical interactions, we define a library $\mathcal{C}$ of 11 reusable primitives (e.g., \textit{Grasp}, \textit{Push}, \textit{Twist}), categorized into \textbf{Spatial Transport}, \textbf{Contact \& Interaction}, and \textbf{Orientation} (Table~\ref{tab:primitive_library}). This library provides a shared basis for both fine-tuning and execution: (1) \textbf{Fine-tuning}: Tasks are disassembled into sequences $\{p_i\}_{i=1}^N \subset \mathcal{C}$ to provide fine-grained supervision for $\mathcal{L}_{prim}$. (2) \textbf{Inference}: Complex tasks are completed by assembling these learned primitives into a structured execution flow.
By grounding the action space in $\mathcal{C}$, we transform the challenge of mastering infinite unique tasks into the strategic assembly of a finite set of robust physical patterns.

\begin{figure*}[t] %
    \centering
    \includegraphics[width=\textwidth]{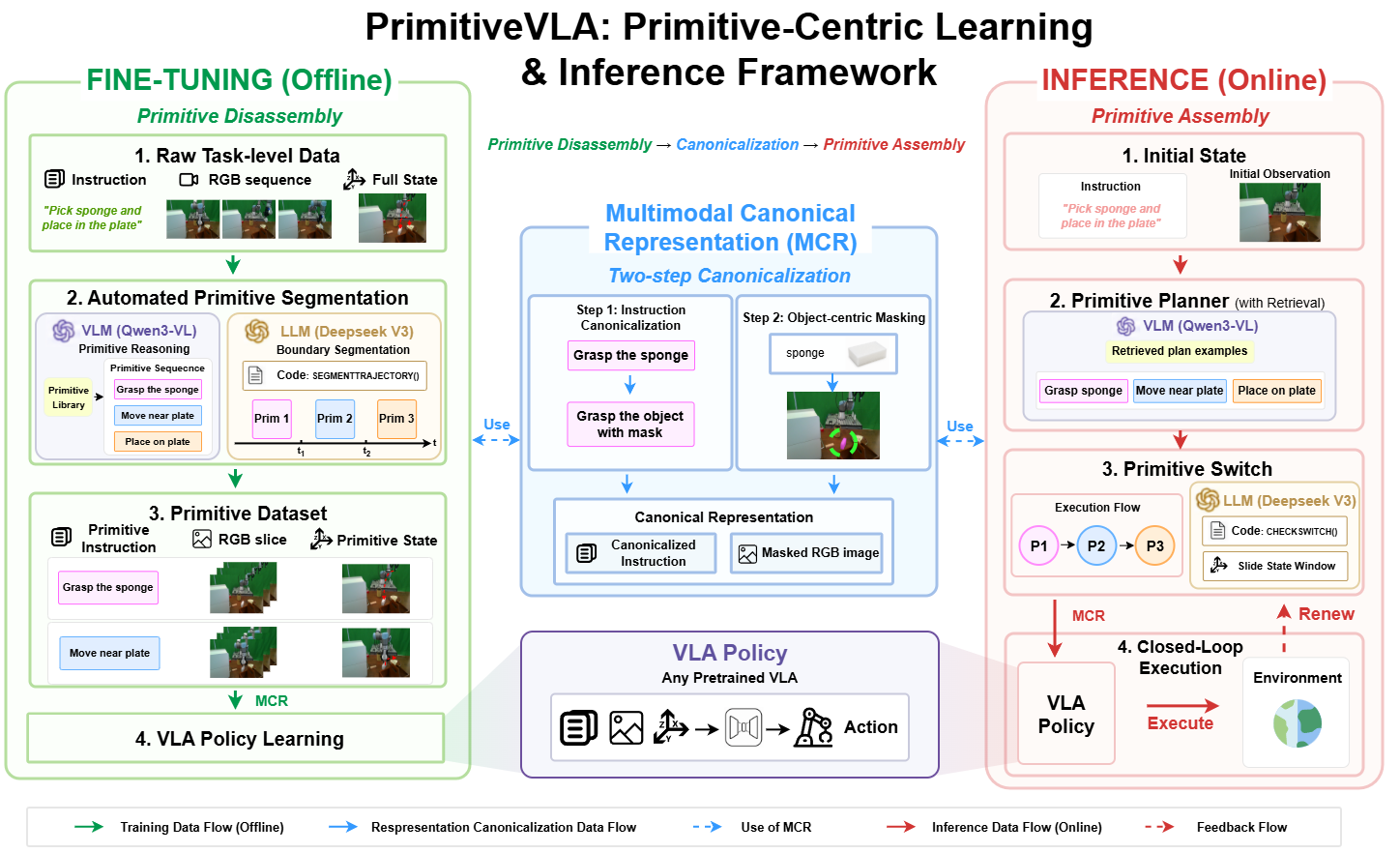}
    \caption{\textbf{Overview of the PrimitiveVLA framework.} The framework models robotic manipulation through reusable primitives, unifying \textbf{fine-tuning} (primitive disassembly) and \textbf{inference} (primitive assembly) via a shared \textbf{Multimodal Canonical Representation} (MCR).}
    \label{fig:overview}
    \vspace{-10pt} 
\end{figure*}

\section{Methodology}
\label{sec:method}

We propose \textbf{PrimitiveVLA}, a Primitive-Centric framework that shifts VLA learning from Direct Instruction-to-Control Mapping to a \textbf{Disassemble \& Assemble paradigm}. As illustrated in Fig.~\ref{fig:overview}, our method consists of two phases: \textbf{Fine-tuning} and \textbf{Inference}.

In the \textbf{Fine-tuning phase} (Sec.~\ref{sec:segmentation}), we \textbf{disassemble} task trajectories into primitives to fine-tune the VLA for mastering physical motion patterns. In the \textbf{Inference phase} (Sec.~\ref{sec:inference}), the VLA \textbf{assembles} these learned behaviors for task execution, guided by a VLM-based planner for primitive selection and a rule-based controller for switching.
To bridge these two phases, we adopt the \textbf{Multimodal Canonical Representation (MCR, Sec.~\ref{sec:representation})}, which unifies inputs into canonical instructions for defining primitives and object-centric masks for preserving task context. This enables MCR to align identical motion patterns across tasks for VLA fine-tuning, while preserving contextual information for improved generalization and reusable primitive assembly during inference.

\subsection{Fine-tuning Phase: Primitive Disassembly}
\label{sec:segmentation}

To enable VLA models to learn reusable motion patterns, fine-grained primitive-level supervision is essential. Since existing datasets predominantly provide monolithic trajectories $\tau$, we propose an \textbf{automated} pipeline to disassemble these into primitive-level data by decoupling the process into two sub-problems: determining the execution structure (\textbf{Primitive Sequence Reasoning}, Sec.~\ref{sec:primitivereasoning}) and localizing their precise boundaries (\textbf{State-Based Boundary Segmentation}, Sec.~\ref{sec:State Segmentation}). The resulting primitive-level data is then used to fine-tune the VLA to master these underlying motion patterns.

\subsubsection{Primitive Sequence Reasoning}
\label{sec:primitivereasoning}
The first step of disassembly is determining the underlying execution structure. We leverage a VLM as a reasoning engine $f_{\text{VLM}}$ to identify the \textbf{primitive sequence} $\mathcal{S} = [p_1, p_2, \dots, p_k]$ for a task:
\begin{equation}
    \mathcal{S} = f_{\text{VLM}}(l, \mathcal{V}_\tau, \mathcal{C}),
\end{equation}
where $l$ is the task instruction, $\mathcal{V}_\tau$ is the example RGB sequence, and $\mathcal{C}$ is the primitive library (in Sec.~\ref{sec:primitive_design}). This sequence $\mathcal{S}$ explicitly defines the temporal order of primitives without boundaries, providing a robust semantic prior for subsequent segmentation. These structured $(l, \mathcal{S})$ pairs are indexed into a \textbf{Disassembly Library} $\mathcal{D}$ to serve as expert exemplars for inference. (in Sec.~\ref{sec:Primitive Planner}).

\subsubsection{State-Based Boundary Segmentation}
\label{sec:State Segmentation}
Once the sequence is established, to localize the start and end points of each primitive, we input the dataset-relevant primitive set $\mathcal{P} = \bigcup_{i} \{p \mid p \in \mathcal{S}_i\}$ (all primitives in the inferred sequences $\{\mathcal{S}_i\}$) and their physical definitions into an LLM to generate \textbf{Python code} $\phi_i$ that defines the termination criteria for each primitive $p_i \in \mathcal{C}$. For a primitive starting at $t_{\text{start}}$, the endpoint $t_{\text{end}}$ is identified as:
\begin{equation}
    t_{\text{end}} = \min \{ t \mid t > t_{\text{start}} + \delta, \,\, \phi_i(s_{t-k:t+k}) = \text{True} \},
\end{equation}
where $\delta$ is a temporal offset and $\phi_i$ evaluates a local window to capture motion dynamics. For example, a \textit{Grasp} primitive is segmented when the gripper width remains constant ($w < \epsilon$) while the end-effector height begins to increase ($\Delta z > \delta$). This transition-based logic provides deterministic physical grounding for disassembly, as detailed in Primitive Segmentation algorithm (Fig.~\ref{fig:Pseudocode} (Left)).

\textbf{Learning with MCR.}
Before fine-tuning, all disassembled segments are processed via MCR to unify cross-modal variance. Specifically, task-specific instructions and raw observations are transformed into canonical instructions $c_i$ and object-centric masks $\tilde{o}_i$ (Sec.~\ref{sec:representation}). By decoupling motion patterns from task-specific context, this transformation aligns identical motion patterns across diverse tasks into a shared primitive. This enables architecture-agnostic learning across tasks, leveraging multi-task fine-tuning to transform the VLA into a primitive master ready for inference-phase assembly.

\begin{figure}[t]
\centering %
\scriptsize 
\renewcommand{\arraystretch}{0.85} 
\setlength{\tabcolsep}{2pt} 

\begin{minipage}[t]{0.35\linewidth} %
\vspace{0pt}
\begin{tabular}{@{}p{\linewidth}@{}}
\toprule
\textbf{Algorithm} Primitive Segmentation \\
\midrule
\textbf{Input:} Trajectory $T$, Primitive $P$, start $t_0$ \\
\textbf{function} \textsc{SegmentTrajectory}$(T, P, t_0)$ \\
\hspace*{0.5em} \textbf{if} $P$ is last \textbf{then} \textbf{return} $|T| - 1$ \\
\hspace*{0.5em} $\Phi \gets \textsc{GetRule}(P)$, $\delta \gets 10$ \\
\hspace*{0.5em} \textbf{for} $t = t_0 + \delta$ \textbf{to} $|T|$ \textbf{do} \\
\hspace*{1.5em} \textbf{if} $\Phi(T, t)$ \textbf{then} \textbf{return} $t$ \\
\hspace*{0.5em} \textbf{end for} \\
\hspace*{0.5em} \textbf{return} Null \\
\bottomrule
\end{tabular}%
\end{minipage}%
\hspace{0.06\linewidth}%
\begin{minipage}[t]{0.35\linewidth}
\vspace{0pt}
\begin{tabular}{@{}p{\linewidth}@{}}
\toprule
\textbf{Algorithm 2:} Primitive Switch \\
\midrule
\textbf{Input:} History Window $\mathcal{H}_t$, Primitive $P$ \\
\textbf{function} \textsc{CheckSwitch}$(\mathcal{H}_t, P)$ \\
\hspace*{0.5em} \textbf{if} $|\mathcal{H}_t| <$ WindowSize \textbf{then} \textbf{return} False \\
\hspace*{0.5em} $s_{\text{stat}} \gets \textsc{CalcStatistics}(\mathcal{H}_t)$ \\
\hspace*{0.5em} $\Phi \gets \textsc{GetCondition}(P)$ \\
\hspace*{0.5em} \textbf{if} $\Phi(s_{\text{stat}})$ \textbf{then} \textbf{return} True \\
\hspace*{0.5em} \textbf{end if} \\
\hspace*{0.5em} \textbf{return} False \\
\bottomrule
\end{tabular}%
\end{minipage}

\caption{Primitive segmentation and switch algorithms.}
\label{fig:Pseudocode}
\vspace{-10pt} 
\end{figure}

\subsection{Inference Phase: Primitive Assembly}
\label{sec:inference}

To bridge the gap between isolated primitives and complex tasks, we design a Primitive Assembly mechanism that orchestrates execution via a retrieval-augmented planner (Sec.~\ref{sec:Primitive Planner}) and a closed-loop switch controller (Sec.~\ref{sec:Primitive Planner}). Finally, the VLA policy executes actions while the primitive switch controller governs transitions, enabling closed-loop execution.

\subsubsection{Primitive Planner}
\label{sec:Primitive Planner}
Inference begins by determining the primitive sequence for task $l_{\text{test}}$. For $l_{\text{test}}$, we retrieve the top-3 similar $(l, \mathcal{S})$ pairs from the Disassembly Library $\mathcal{D}$ via semantic cosine similarity. These priors, along with the instruction $l_{\text{test}}$, initial observation $o_0$, and primitive definitions $\mathcal{C}$, are fed into the VLM to generate the sequence:
\begin{equation}
    \mathcal{S}_{\text{plan}} = f_{\text{VLM}}(l_{\text{test}}, o_0, \mathcal{C}, \text{Retrieve}(l_{\text{test}}, \mathcal{D})).
\end{equation}
Through In-Context Learning, this constrains the VLM to generate sequences that adhere to the fine-tuning distribution, preventing the hallucination of unreasonable primitive combinations.

\subsubsection{Primitive Switch}
\label{sec:Primitive Switch}
Successful multi-stage assembly hinges on robust switching. By inputting the task-relevant primitive set $\mathcal{P}_{\text{test}} = \bigcup_{i} \{p \mid p \in \mathcal{S}_{plan}^i\}$, physical definitions, and the previously generated segmentation code $\phi_i$, we leverage an LLM to generate Python code $f_{\text{switch}}$. This code mirror the segmentation criteria $\phi_i$ but are optimized to monitor statistical trends $s_{\text{stat}}$ (e.g., rate of change) from a \textbf{sliding history window} $\mathcal{H}_t$ of size $W$. As shown in Fig.~\ref{fig:Pseudocode} (Right), a transition from $p_i$ to $p_{i+1}$ is triggered when:
\begin{equation}
\text{SwitchTrigger} \iff \forall s \in s_{\text{stat}}, \quad f_{\text{switch}}(s, p_i) = \text{True}.
\end{equation}
By evaluating consecutive frames within the window, $f_{\text{switch}}$ effectively mitigates sensor noise and prevents premature transitions. During online execution, the system maintains a \textbf{dual-threaded architecture}: the execution thread outputs actions via the VLA model, while the monitoring thread concurrently evaluates the switch trigger. This ensures task progression through the compositional \textbf{assembly} of primitives, realizing a reactive closed-loop execution. During each control step, the live observation and the planned primitive $p_i$ are processed by \textbf{MCR} to maintain consistency with the fine-tuning distribution, ensuring the VLA accurately synthesizes actions.

\subsection{Multimodal Canonical Representation (MCR)}
\label{sec:representation}

To mitigate \textbf{contextual interference}, MCR transforms multimodal inputs into a standardized canonical space. By unifying instructions and utilizing masks to carry task-specific information, MCR ensures that inputs from diverse tasks maintain a consistent feature property. This allows the VLA to focus on learning universal motion patterns through a unified interface, while the mask provides the necessary context for precise control. As a result, MCR aligns observations into a compatible representation, ensuring that learned primitives remain robust and transferable across environments.

\noindent \textbf{Semantic Unification.} We map all samples of a primitive $p_i$ to a single \textbf{canonical primitive instruction} $c_i$. For instance, task-specific instructions like \textit{"grasp the black bowl"} are unified into \textit{"grasp the object with green mask"}. By constraining the language space, we guide the model to associate policies with shared primitive identities rather than memorizing textual variations.

\noindent \textbf{Visual Compatibility.} We handle visual variations using \textbf{masked observations} $\tilde{o}_i = o_i \otimes M_i$, where $M_i$ is an object-centric mask tracked via SAM~\cite{kirillov2023segany} and Cutie~\cite{cheng2024putting}. This mechanism ensures that VLA inputs maintain a uniform representation across tasks, while the masks serve as the primary carrier of task-specific context. By applying these masks, we maintain a uniform feature format for the VLA to learn primitives, while allowing the model to understand the task environment through the masked visual information.

\section{Experiments}
\label{sec:experiments}

\looseness=-1 In the preceding section, we established the \textbf{PrimitiveVLA} framework to address the data efficiency and generalization bottlenecks of generalist VLA models. In this section, we validate the effectiveness of this framework across comprehensive simulation and real-world experiments to address three core Research Questions (RQs):
\vspace{-5pt}
\begin{itemize}
    \item \looseness=-1 \textbf{RQ1 (Performance \& Efficiency):} Can the VLA models fine-tuned via our framework demonstrate superior \textbf{data efficiency} compared to standard VLAs under identical data conditions, and \textbf{match the full-data baseline performance} even with \textbf{50\%} training data?
    
    \item \textbf{RQ2 (Generalization Capabilities):} Can the VLA models fine-tuned via our framework effectively generalize to \textbf{unseen tasks} (out-of-domain) and \textbf{long-horizon tasks}?
    
    \item \textbf{RQ3 (Mechanism Verification):} What are the individual contributions of "Primitive Disassembly" and "MCR Representation" to the overall performance gain?
\end{itemize}
\vspace{-10pt}
\subsection{Experimental Setup}
\label{sec:setup}

\vspace{-5pt}
\looseness=-1 \noindent \textbf{Simulation Benchmarks.} We primarily utilize \textbf{Libero}~\cite{liu2023libero}: for fine-tuning/in-domain (ID) task, we use \textbf{Libero-Spatial}, \textbf{Object}, \textbf{Goal} (10 tasks each), plus a \textbf{Libero-90} subset (84 tasks, excluding 6 low-quality ones); 
for out-of-domain (OOD) Evaluation, We design a \textbf{custom} \textbf{Libero-90-Novel} suite (8 tasks with unseen logic, listed in Table~\ref{tab:libero_novel}) and \textbf{Libero-Long} to test generalization.
Additionally, to stress-test robustness, we evaluate on \textbf{RLBench}~\cite{james2020rlbench} using 10 tasks characterized by higher data diversity: \textit{Meat Off Grill, Lamp Off, Push Button, Open Jar, Lift Numbered Block, Open Wine Bottle, Take Off Scales, Take Lid Off Saucepan, Put Rubbish In Bin, Take Umbrella}.

\begin{table}[tbh]
    \centering
    \footnotesize
    \vspace{-10pt}
    \caption{\textbf{Libero-90-Novel Tasks.} We evaluate generalization on 8 customized unseen tasks.}
    \vspace{1em}
    \label{tab:libero_novel}
    \resizebox{\linewidth}{!}{%
    \begin{tabular}{@{}ll ll@{}}
        \toprule
        \textbf{Scene} & \textbf{Instruction} & \textbf{Scene} & \textbf{Instruction} \\
        \midrule
        Kit-1 & Open top drawer, place bowl in, close & Liv-4 & Pick right black bowl, place in tray \\
        Kit-2 & Stack front black bowl on back & Liv-5 & Place yellow/white mug on left plate \\
        Kit-10 & Open top drawer, place pudding in, close & Liv-2 & Pick butter, place in basket \\
        Kit-10 & Close middle cabinet drawer & Kit-10 & Place black bowl on cabinet top \\
        \bottomrule
    \end{tabular}%
    }
    \vspace{-5pt}
\end{table}

\looseness=-1 \noindent \textbf{Real-World Environment.} To validate physical transferability, we use a UR5e arm with a Robotiq 2F-85 gripper and two RealSense L515 cameras (one third-person and one wrist-mounted, shown in \textbf{Appendix}). We evaluate \textbf{11 tasks}:
(1) ID: \textit{Pick sponge on plate; Pick cup on cabinet; Pick cube in top drawer; Pick cube in caddy; Push top drawer; Push bottom drawer}.
(2) Task Generalization: \textit{Pick sponge in caddy (unseen pair); Pick cube on cabinet (unseen pair); Pick blue cup on cabinet (novel object)}.
(3) Compositional Generalization: \textit{Pick cup then push drawer; Pick sponge then cube}.

\noindent \textbf{Base Models.} To demonstrate that PrimitiveVLA decouples the underlying VLA backbone, we validate it across the full spectrum of VLA paradigms:
\textbf{OpenVLA}~\cite{kim2024openvla} (Standard Transformer-based VLA);
\textbf{OpenVLA-OFT}~\cite{kim2025optimizing} (Optimized Fine-Tuning variant of OpenVLA); and
\textbf{${\pi_{0.5}}$}~\cite{intelligence2025pi05} (SOTA foundation model utilizing flow matching). 

\noindent \textbf{Metrics.} We report the average \textbf{Success Rate}. Simulation tasks are evaluated over 50 trials, while Real-World and RLBench tasks are evaluated over 20 trials.

\noindent \textbf{More details are listed in Appendix.}

\begin{table}[t]
    \vspace{-10pt}
    \centering
    \scriptsize
    \caption{\textbf{Libero Evaluation.} We report performance on Libero-Object, Libero-Spatial, Libero-Goal and Libero-90, also out-of-distribution 0-shot generalization on Libero-90-Novel and Libero-Long. 
    The input modalities include language instructions (L), global observations ($I_g$), wrist observations ($I_w$), and proprioceptive states (S). ``-'' indicates unavailable results.}
    \label{tab:unified_results}
    \vspace{1em}
    \renewcommand{\arraystretch}{0.9}

    \begin{tabular}{l l cccc c cc}
        \toprule
        & & \multicolumn{4}{c}{\textbf{Small-Scale}} 
        & \textbf{Large-Scale} 
        & \multicolumn{2}{c}{\textbf{0-shot Generalization}} \\
        \cmidrule(lr){3-6} \cmidrule(lr){7-7} \cmidrule(lr){8-9}
        \textbf{Model} & \textbf{Input Modality} 
        & Object & Spatial & Goal & \textbf{Mean} 
        & 90 
        & 90-Novel & Long \\
        \midrule
        
        TraceVLA~\cite{tracevla2025}       
        & $L, I_{g}$ 
        & 84.60\% & 85.20\% & 75.10\% & 81.63\% 
        & - & - & - \\

        SpatialVLA~\cite{spatialvla2025}   
        & $L, I_{g}$ 
        & 88.20\% & 89.90\% & 78.60\% & 85.57\% 
        & - & - & - \\

        OpenVLA~\cite{kim2024openvla}      
        & $L, I_{g}$ 
        & 87.40\% & 82.80\% & 74.00\% & 82.40\% 
        & 70.60\% & 7.38\% & 4.50\% \\

        \textbf{OpenVLA + Ours}            
        & $L, I_{g}$ 
        & \textbf{90.60\%} & \textbf{91.20\%} & \textbf{82.20\%} & \textbf{88.00\%} 
        & \textbf{79.80\%} & \textbf{45.50\%} & \textbf{38.50\%} \\

        \midrule
        
        Diffusion Policy~\cite{chi2023diffusion} 
        & $L, I_{g}, I_{w}, S$ 
        & 92.50\% & 78.30\% & 68.30\% & 79.70\% 
        & - & - & - \\

        Octo~\cite{octo_2023}                    
        & $L, I_{g}, I_{w}, S$ 
        & 85.70\% & 78.90\% & 84.60\% & 83.07\% 
        & - & - & - \\

        $\pi_{0}$~\cite{black2024pi0}            
        & $L, I_{g}, I_{w}, S$ 
        & 96.80\% & 98.80\% & 95.80\% & 97.13\% 
        & - & - & - \\

        OpenVLA-OFT~\cite{kim2025optimizing}     
        & $L, I_{g}, I_{w}, S$ 
        & 97.40\% & 98.60\% & 96.80\% & 97.60\% 
        & 89.70\% & 13.50\% & 3.75\% \\

        \textbf{OpenVLA-OFT + Ours}              
        & $L, I_{g}, I_{w}, S$ 
        & \textbf{98.40\%} & \textbf{99.40\%} & \textbf{97.80\%} & \textbf{98.53\%} 
        & \textbf{94.70\%} & \textbf{71.00\%} & \textbf{66.50\%} \\
        
        $\pi_{0.5}$~\cite{intelligence2025pi05} 
        & $L, I_{g}, I_{w}, S$ 
        & 98.20\% & 98.40\% & 96.00\% & 97.53\% 
        & 96.80\% & 56.00\% & 30.50\% \\

        \textbf{$\pi_{0.5}$ + Ours}             
        & $L, I_{g}, I_{w}, S$ 
        & 98.00\% & 98.20\% & 95.80\% & 97.33\% 
        & 96.20\% & \textbf{75.50\%} & \textbf{80.25\%} \\

        \bottomrule
    \end{tabular}

    \vspace{-5pt}
\end{table}

\vspace{-5pt}
\subsection{Data Efficiency (RQ1)}
\label{sec:performance_efficiency}

To answer \textbf{RQ1}, we evaluate the data efficiency of PrimitiveVLA. We benchmark diverse architectures on \textbf{Libero} under the full-data regime and specifically assess performance under a \textbf{half-scale} (50\% data) setting. Additionally, we extend the evaluation to the complex \textbf{RLBench} suite to verify robustness in high-variance environments.

\noindent \textbf{Superiority under Full-Data Regime.} PrimitiveVLA consistently enhances or matches baselines across backbones (Table~\ref{tab:unified_results}). On the large-scale \textbf{Libero-90} (Table~\ref{tab:unified_results}), our method boosts the OpenVLA baseline by \textbf{9.2\%} and OpenVLA-OFT by \textbf{5.0\%}. On the Small-Scale benchmark, PrimitiveVLA also yields substantial improvements for both OpenVLA (+5.6\%) and OpenVLA-OFT. Even when applied to the state-of-the-art $\pi_{0.5}$, our method maintains its competitive performance. Furthermore, on the complex \textbf{RLBench} suite (Table~\ref{tab:rlbench}), PrimitiveVLA improves the average success rate by \textbf{7.0\%}.

\vspace{-3pt}
\noindent \textbf{Matching Full-Data Performance with 50\% Data.} Remarkably, our method maintains high performance even when training data is drastically reduced. As shown in Table~\ref{tab:low_data}, OpenVLA-based PrimitiveVLA with \textbf{50\% data} (80.30\%) outperforms the 100\%-data OpenVLA baseline. For OFT, the 50\%-data variant even leads its 100\%-data baseline by \textbf{2.4\%} on Libero-90. These results demonstrate that our paradigm enables the VLA to effectively learn shared motion patterns across diverse tasks, facilitating high performance even in data-constrained scenarios.

\begin{table}[t]
    \vspace{-5pt}
    \centering
    \scriptsize
    \caption{\textbf{Data Efficiency Analysis.} We compare four settings across different VLA backbones, varying both the training data ratio (50\% vs. 100\%) and the method (baseline vs. PrimitiveVLA).}
    \label{tab:low_data}
    \vspace{1em}
    
    \setlength{\tabcolsep}{3pt}
    \renewcommand{\arraystretch}{0.9}
    
    \begin{tabular}{lc cccc c}
        \toprule
        &  & \multicolumn{3}{c}{\textbf{Small-Scale}} & \textbf{Large-Scale}  \\
        \cmidrule(lr){3-5} \cmidrule(lr){6-6}
        \textbf{Model} & \textbf{Data Scale} 
        & Libero-Object & Libero-Spatial & Libero-Goal
        & Libero-90 & \textbf{Mean} \\
        \midrule
        
        \multirow{2}{*}{OpenVLA (Base)} 
        & 50\% 
        & 83.40\% & 73.40\% & 73.60\%  
        & 65.89\% & 74.07\% \\
        & 100\% 
        & 87.40\% & 82.80\% & 74.00\% 
        & 70.60\% & 78.70\%  \\
        \cmidrule(lr{1em}){1-7} %
        \multirow{2}{*}{OpenVLA + Ours}
        & 50\%
        & 87.60\% & 87.00\% & 73.20\% 
        & 73.40\% & 80.30\%  \\
        & 100\%
        & \textbf{90.60\%} & \textbf{91.20\%} & \textbf{82.20}\% 
        & \textbf{79.80\%} & \textbf{85.95\%}  \\

        \specialrule{0.08em}{0.4em}{0.4em} %

        \multirow{2}{*}{OpenVLA-OFT (Base)} 
        & 50\% 
        & 96.20\% & 97.00\% & 91.00\%  
        & 87.00\% & 92.80\% \\
        & 100\% 
        & 97.40\% & 98.60\% & 96.80\% 
        & 89.70\% & 95.63\%  \\
        \cmidrule(lr{1em}){1-7} %
        \multirow{2}{*}{OpenVLA-OFT + Ours}
        & 50\%
        & 94.80\% & 97.80\% & 96.00\% 
        & 92.12\% & 95.18\%  \\
        & 100\%
        & \textbf{98.40\%} & \textbf{99.40\%} & \textbf{97.80}\% 
        & \textbf{94.70\%} & \textbf{97.58\%}  \\
        
        \bottomrule
    \end{tabular}
    
    \vspace{-15pt}
\end{table}

\begin{table}[t]
    \vspace{-5pt}
    \centering
    \scriptsize
    \caption{\textbf{Success Rate on RLBench.} We evaluate the 10 manipulation tasks detailed in Sec.~\ref{sec:setup} (denoted as T1--T10). PrimitiveVLA achieves a \textbf{7.0\% improvement} over the baseline.}
    \label{tab:rlbench}
    \vspace{1em}
    \renewcommand{\arraystretch}{0.9}
    \begin{tabular}{l rrrrrrrrrr r}
        \toprule
        \textbf{Model} & T1 & T2 & T3 & T4 & T5 & T6 & T7 & T8 & T9 & T10 & \textbf{Mean} \\
        \midrule
        OpenVLA & 25\% & 90\% & 95\% & 20\% & 40\% & 20\% & 15\% & 95\% & 40\% & 55\% & 49.5\% \\
        \textbf{OpenVLA + Ours} & \textbf{35\%} & \textbf{100\%} & 95\% & 15\% & 35\% & \textbf{30\%} & \textbf{40\%} & \textbf{100\%} & \textbf{50\%} & \textbf{65\%} & \textbf{56.5\%} \\
        \bottomrule
    \end{tabular}%
    
    \vspace{-5pt}
\end{table}

\vspace{-5pt}
\subsection{Cross-Task Generalization (RQ2)}
\label{sec:generalization}
\vspace{-5pt}
To answer \textbf{RQ2}, we investigate whether the VLA fine-tuned via PrimitiveVLA can generalize to OOD tasks. We evaluate the \textbf{0-shot performance} of models fine-tuned on \textbf{Libero-90} against two OOD suites: unseen tasks (\textbf{Libero-90-Novel}) and long-horizon sequences (\textbf{Libero-Long}) (Table~\ref{tab:unified_results} (Right)). The evaluation configurations remain strictly consistent with the settings used for the Libero-90.

\noindent \textbf{Generalization to Unseen Tasks.} Standard VLAs show limited zero-shot generalization on Libero-90-Novel: OpenVLA and OFT achieve only 7.38\% and 13.50\%, while the SOTA $\pi_{0.5}$ reaches 56.00\%. In contrast, PrimitiveVLA elevates these scores to \textbf{45.50\%} (\textbf{6$\times$ improvement} for OpenVLA), \textbf{71.00\%}, and \textbf{75.50\%}. This validates that disassembling tasks into reusable primitives enables the flexible assembly of learned primitives for novel tasks, avoiding task-level overfitting.

\vspace{-3pt}
\noindent \textbf{Generalization to Long-Horizon Tasks.} We evaluate Libero-Long using a reset protocol, where the robot resets to initial pose after each sub-task to mitigate OOD drift at transition gaps.
While standard baselines fail catastrophically (success rates $<$ 5\%), PrimitiveVLA delivers consistent gains, notably boosting the SOTA $\pi_{0.5}$ from 30.50\% to \textbf{80.25\%}.
This result confirms that our assembly-based method provides robust sequential generalization where direct mapping models struggle.

\subsection{Ablation Studies (RQ3)}
\label{sec:ablation}

\vspace{-5pt}
To answer \textbf{RQ3}, we decouple the contributions of Primitive Disassembly and Multimodal Canonical Representation (MCR) by comparing ablated variants on Libero-90 (ID), Novel (OOD), and Long (Long-Horizon) tasks (Table~\ref{tab:ablation}). 

\noindent MCR is the primary driver for OOD transfer, improving OFT's Libero-90-Novel success from 13.50\% to \textbf{60.00\%}. Conversely, Primitive Disassembly is decisive for long-horizon stability, boosting success on \textbf{Libero-Long} from 3.75\% to \textbf{52.30\%} (OFT). Collectively, these results confirm that MCR provides environmental robustness for novel tasks, while Primitive Disassembly ensures the execution stability required for complex, multi-stage execution.

\begin{table}[t]
    \vspace{-5pt}
    \centering
    \scriptsize
    \caption{\textbf{Ablation Study.} We decouple component contributions across ID, OOD, and Long-Horizon tasks. Results demonstrate that \textbf{MCR} is the primary driver for generalization, while \textbf{Primitive Disassembly} ensures temporal consistency.}
    \label{tab:ablation}
    \vspace{1em}
    \setlength{\tabcolsep}{4pt}        %
    \renewcommand{\arraystretch}{0.9}  %
    
    \begin{tabular}{llccc}
        \toprule
        \textbf{Base Model} & \textbf{Setting} & \textbf{Libero-90 (ID)} & \textbf{Libero-90-Novel (OOD)} & \textbf{Libero-Long} \\
        \midrule
        \multirow{4}{*}{OpenVLA} 
          & Baseline & 70.60\% & 7.38\% & 4.50\% \\
          & w/o MCR (Disass. Only) & 73.90\% & 21.50\% & 28.20\% \\
          & w/o Disass. (MCR Only) & 79.60\% & 33.75\% & 4.25\% \\
          & \textbf{Ours (PrimitiveVLA)} & \textbf{79.80\%} & \textbf{45.50\%} & \textbf{38.50\%} \\
        \midrule
        \multirow{4}{*}{OpenVLA-OFT} 
          & Baseline & 89.70\% & 13.50\% & 3.75\% \\
          & w/o MCR (Disass. Only) & 89.60\% & 15.00\% & 52.30\% \\
          & w/o Disass. (MCR Only) & 94.30\% & 60.00\% & 39.75\% \\
          & \textbf{Ours (PrimitiveVLA)} & \textbf{94.70\%} & \textbf{71.00\%} & \textbf{66.50\%} \\
        \bottomrule
    \end{tabular}
    
    \vspace{-15pt}
\end{table}

\begin{table}[tbh]
    \vspace{-10pt}
    \centering
    \scriptsize
    \caption{\textbf{Real-World Evaluation Results.} We report real-world success rates for In-Distribution (T1--T6), Task Generalization (T7--T9), and Compositional Generalization (T10--T11).}
    \label{tab:real_world}
    \vspace{1em}
    \renewcommand{\arraystretch}{0.9}
    \setlength{\tabcolsep}{2.5pt} %
    \begin{tabular}{l cccccc c ccc c cc c}
    \toprule
    & \multicolumn{7}{c}{\textbf{In-Distribution (ID)}} & \multicolumn{4}{c}{\textbf{Task Gen. (OOD)}} & \multicolumn{3}{c}{\textbf{Compositional}} \\
    \cmidrule(lr){2-8} \cmidrule(lr){9-12} \cmidrule(lr){13-15}
    \textbf{Model} & T1 & T2 & T3 & T4 & T5 & T6 & \textbf{Mean} & T7 & T8 & T9 & \textbf{Mean} & T10 & T11 & \textbf{Mean} \\
    \midrule
    $\pi_{0.5}$ & 60\% & 75\% & 60\% & 45\% & 95\% & 85\% & 70\% & 15\% & 35\% & 10\% & 20\% & 0\% & 20\% & 10\% \\
    \textbf{$\pi_{0.5}$ + Ours} & \textbf{85\%} & \textbf{95\%} & \textbf{85\%} & \textbf{85\%} & \textbf{100\%} & \textbf{90\%} & \textbf{90\%} & \textbf{40\%} & \textbf{50\%} & \textbf{80\%} & \textbf{57\%} & \textbf{60\%} & \textbf{70\%} & \textbf{65\%} \\
    \bottomrule
\end{tabular}
    \vspace{-10pt}
\end{table}

\subsection{Real-World Evaluation}
\label{sec:real_world}

\vspace{-5pt}
We deployed \textbf{PrimitiveVLA} on a UR5e robot using the SOTA ${\pi_{0.5}}$ backbone. We collected demonstrations with broad spatial variance and evaluated 11 tasks (T1--T11) covering In-Distribution (ID), Task Generalization, and Compositional Generalization settings (Table~\ref{tab:real_world}).

\noindent In ID settings (T1--T6), our method achieves an \textbf{90\%} average success rate, significantly outperforming the ${\pi_{0.5}}$ baseline (70.0\%) and demonstrating superior robustness against physical variance. Our framework further enables remarkable 0-shot generalization: it boosts success on unseen objects (T9) from 10\% to \textbf{80\%} and excels in compositional tasks where the baseline fails (T10: 0\%), achieving \textbf{60\%} and \textbf{70\%} success on T10 and T11, respectively. These results confirm that our reusable primitives effectively enables the zero-shot assembly in novel, long-horizon real-world scenarios.

\vspace{-5pt}
\section{Conclusion}
\label{sec:Conclusion}
\looseness=-1 We propose \textbf{PrimitiveVLA} to resolve the dual bottlenecks of \textbf{data inefficiency} and \textbf{limited generalization} in current VLAs by operationalizing a \textbf{disassemble-and-assemble} paradigm. By unifying \textbf{fine-tuning-phase disassembly} and \textbf{inference-phase assembly} through a shared \textbf{Multimodal Canonical Representation (MCR)}, our framework shifts the paradigm from memorizing monolithic trajectories to mastering a set of reusable physical primitives. Experiments confirm that PrimitiveVLA matches or exceeds baselines performance using only \textbf{50\% data} and enables robust \textbf{zero-shot assembly} for unseen and long-horizon tasks, demonstrating that the \textbf{disassembly} of physical interactions into reusable primitives is essential for scalable Embodied AI.

\noindent \textbf{Limitations and Future Work.} Currently, our primitive set relies on a pre-defined kinematic taxonomy. While this set is comprehensive for standard gripper-based manipulation, it may not cover highly specialized dexterous manipulations. Future work will explore \textbf{unsupervised primitive disassembly} to dynamically expand the action space and investigate \textbf{end-to-end differentiable planning} to further optimize the synergy between high-level reasoning and low-level control.

\newpage{}
\bibliographystyle{unsrtnat}
{
\small
\bibliography{nips}
}

\newpage{}

\appendix
\section*{\Large Appendix}
\section{More Details about Method}
\subsection{Fine-tuning: Primitive Reasoning}
To ensure structural coherence during the disassembly process, we employ a VLM-based semantic decomposer as the primary step to establish a high-level task framework. While it is technically feasible to segment trajectories using purely state-based code, such a non-prior approach is often hypersensitive to minor execution drifts in unstructured or human-demonstrated trajectories.

Specifically, during a manual "move" operation, incidental axial shifts or unintended tremors may appear as prominent deviations in the robotic proprioceptive data (e.g., a slight rotation or height change). Without a semantic prior, these jitters might be incorrectly segmented into redundant, fine-grained primitives (e.g., an independent "Rotate" or "Adjust" phase). However, from a visual and semantic perspective, these deviations are often negligible and should be subsumed within the primary "Move" primitive.

By leveraging the visual understanding and logical reasoning of VLMs, PrimitiveVLA identifies the macroscopic task flow first. This top-down guidance filters out numerical noise in the trajectory that does not align with the intended semantic action, preventing over-segmentation and ensuring that the resulting primitives are both physically grounded and semantically meaningful.

To facilitate structured disassembly, we feed the VLM with a comprehensive input tuple: the high-level task description, the sampled RGB frame sequence of the demonstration, and our predefined library of physical primitives. By grounding the visual observations in these semantic and physical constraints, the VLM generates a reasoned primitive sequence for each task.

The reasoned outputs from the VLM are not merely transient labels; they serve as the foundational knowledge for our execution framework. For each task, the VLM outputs a Instruction-Sequence Pair $(Inst, Seq)$, where $Inst$ represents the natural language of the task and $Seq$ is the corresponding physical primitive sequence from VLM. These pairs are systematically archived into a Disassembly Library $\mathcal{D}$. During the Inference Phase, this library acts as a structured memory for the High-Level Planner. When a new task is encountered, the planner queries the Disassembly Library to retrieve and adapt successful primitive strategies.

Specifically, we adopt \textbf{Qwen3-VL}~\cite{qwen3vl2025} as the VLM in our framework, due to its strong video understanding and logical reasoning capabilities, as well as its suitability as an open-source model. Here is an illustrative example of the prompt used to guide the VLM in Primitive Reasoning to disassemble unstructured demonstrations:

\newtcolorbox{promptbox}[1][]{
    colback=gray!10, %
    colframe=gray!50, %
    arc=2pt,         %
    boxrule=0.8pt,   %
    left=10pt, right=10pt, top=10pt, bottom=10pt,
    fontupper=\small\ttfamily, %
    title=#1,
    coltitle=black,
    fonttitle=\bfseries\sffamily,
    attach title to upper=\quad,
    after skip=10pt, before skip=10pt,
    breakable
}

\begin{promptbox}[System Prompt: Primitive Reasoning] \\
\textbf{[System Role]} \\
You are a robotic reasoning expert. Your goal is to decompose a long-horizon manipulation video into a structured sequence of physical primitives based on the provided library and merging rules.

\vspace{5pt}
\textbf{[Primitive Definitions]}
\begin{enumerate}[leftmargin=1.5em, itemsep=0pt]
    \item \textbf{grasp}: Move toward an object, seize it, and lift slightly.
    \item \textbf{lift}: Move upward significantly ($z$-axis) with closed gripper to reach a height, possibly with minor $xy$ adjustment.
    \item \textbf{move}: Large-scale movement in the $xy$-plane while the gripper is closed.
    \item \textbf{place}: Descend to release an object at target location and lift away.
    \item \textbf{push}: Apply force to slide an object to a target position.
    \item \textbf{press}: Apply downward force on an object or a surface.
    \item \textbf{twist}: Rotate the gripper around its central axis (e.g., turning a knob).
    \item \textbf{tilt}: Rotate around a joint beyond the wrist (adduction/abduction of the arm).
    \item \textbf{rotate}: Rotate along a fixed axis of an object (e.g., opening a laptop).
    \item \textbf{pull}: Pull an object toward a target position.
    \item \textbf{insert}: Insert an object or the gripper into a gap/slot.
\end{enumerate}

\vspace{5pt}
\textbf{[Merging Rules]} \\
\begin{itemize}[leftmargin=1.2em, label=$\bullet$, itemsep=0pt]
    \item \textbf{Object-Independent Pre-movement:} Motions without object interaction (e.g., approaching a mug) should be merged into the subsequent primitive (e.g., use \textit{grasp} instead of \textit{move + grasp}).
    \item \textbf{State Maintenance:} If the arm maintains a grip, label as \textit{move} unless it aligns with \textit{lift} or \textit{push}. Do not fragment unintentional drifts or jitter.
\end{itemize}

\vspace{5pt}
\textbf{[Input]} \\
Task Instruction: "\{task\}" \\
Video Observation: [Sequence of visual frames $I_{1:T}$]

\vspace{5pt}
\textbf{[Output Requirement]} \\
Reasoning: [Explain the flow and why certain drifts were merged.] \\
Sequence: [Primitive\_1(Instruction), Primitive\_2(Instruction), ...]
\end{promptbox}

\subsection{Fine-tuning: Boundary Segmentation}
While we employ a VLM to extract the high-level task framework, we deliberately avoid using it as the final segmentation criterion. As analyzed in Appendix \ref{appendix:Comparison of Disassembly}, the inherent "black-box" nature and stochasticity of VLMs often lead to inconsistent temporal boundaries. To ensure the precision and stability required for robotic data, we adopt a \textbf{rule-based} execution approach.

However, manually authoring segmentation code for each specific task is as labor-intensive as designing individual reward functions. To overcome this, we utilize an LLM-driven automated code generation pipeline. Our core philosophy is to move beyond task-specific scripts toward a \textbf{unified segmentation standard} applicable across the entire dataset. By providing the LLM with the full spectrum of primitive types (derived from Primitive Reasoning), their semantic definitions, and generic segmentation criteria, we generate code capable of extracting reusable physical primitives rather than task-specialized sub-tasks.

To ensure robustness, the system incorporates the following design considerations:

\begin{itemize}
    \item \textbf{Temporal Window Consistency:} Rather than relying on a single time step, each primitive's termination is determined by analyzing a local window (the states before and after a candidate moment). This prevents premature triggers caused by sensor noise and ensures a deterministic physical grounding.
    
    \item \textbf{Optional Threshold Calibration:} For complex scenarios, the prompt can be optionally augmented with a few segmentation exemplars. This allows the LLM to more accurately calibrate numerical thresholds for specific primitive transitions when necessary.
\end{itemize}

Specifically, we adopt \textbf{DeepSeek-V3}~\cite{deepseekv32024} as the LLM in our framework due to its strong logical reasoning and code generation capabilities, as well as its suitability as an open-weight model. Here is an illustrative example of the prompt used to guide the LLM in generating \textbf{boundary segmentation code} to determine the start and end of each primitive:

\begin{promptbox}[System Prompt: Code Generation for Primitive Segmentation] \\
\textbf{[System Role]} \\
You are an expert AI programming assistant. Generate a Python function to determine the termination index of robotic primitives based on the provided physical criteria and trajectory examples.

\vspace{5pt}
\textbf{[Primitive Segmentation Criteria]}
\begin{enumerate}[leftmargin=1.5em, itemsep=2pt]
    \item \textbf{grasp}: Gripper state remains constant (closed) for current and future frames; action is "close" (1); terminates when future frames show significant movement in $xy$ or $z$.
    \item \textbf{lift}: Significant $z$-axis increase over past frames; terminates when upward motion stops, begins descending, or $xy$ motion ceases.
    \item \textbf{move}: Large $xy$ displacement over past frames; terminates when $xy$ motion stops or $z$ begins a sharp descent.
    \item \textbf{place}: Gripper state is "open" (0) for current and past frames; terminates when a $z$-axis lift (departure) is detected.
    \item \textbf{push}: Significant $xy$ displacement over past frames while $z$ remains stable; terminates when $xy$ motion ceases.
\end{enumerate}

\vspace{5pt}
\textbf{(Optional) [In-Context Exemplar]} \\
\textit{// Example: Segmentation at $t=40$} \\
\textbf{Task}: "Pick up cup" | \textbf{Primitive}: "grasp" | \textbf{Split Index ($t$)}: 40 \\
\begin{center}
\small
\begin{tabular}{c|c|ccc|c|l}
\hline
\textbf{Index} & \textbf{Pos} & \textbf{x} & \textbf{y} & \textbf{z} & \textbf{grip} & \textbf{Window Logic} \\ \hline
35 & $t-5$ & 0.120 & 0.450 & 0.020 & 0.98 & Window Start: Gripper stable \\
... & ... & ... & ... & ... & ... & ... \\
\textbf{40} & \textbf{$t$} & \textbf{0.121} & \textbf{0.450} & \textbf{0.022} & \textbf{0.99} & \textbf{Target Split Point} \\
... & ... & ... & ... & ... & ... & ... \\
45 & $t+5$ & 0.126 & 0.455 & 0.045 & 0.99 & Window End: Sustained $z$-lift \\ \hline
\end{tabular}
\end{center}
\textbf{Split Logic}: At index 40, the criterion is met because for all $i \in [35, 45]$, the gripper is closed, and the velocity trend in this 10-step window shows a transition to vertical rising.

\vspace{5pt}
\textbf{[Functional Requirements]}
\begin{itemize}[leftmargin=1.2em, label=$\bullet$, itemsep=0pt]
    \item \textbf{Input}: \texttt{states} ($T \times 7$), \texttt{actions} ($T \times 7$), \texttt{primitive\_name}, \texttt{t\_start}, \texttt{is\_last}.
    \item \textbf{Temporal Consistency}: Evaluation must start from $t_{\text{start}} + 10$.
    \item \textbf{Window-Based Decision}: Analyze local trends in a window (e.g., 5--10 frames) to ensure physical state changes are sustained and not noise-induced.
\end{itemize}

\vspace{5pt}
\textbf{[Output Format]} \\
Return a standalone Python function: \texttt{def get\_segmentation\_index(...):}
\end{promptbox}

\subsection{Inference: Primitive Planner}
During the inference phase, the robot must generate a primitive execution flow for a target task without the benefit of a pre-recorded demonstration video (RGB sequence). To mitigate the inherent stochasticity and potential "hallucination" of the VLM—which might otherwise suggest unseen primitives or illogical sequences—we implement a \textbf{Retrieval-Augmented Planning} strategy.

We maintain a \textbf{Disassembly Library} $\mathcal{D}$ containing successful $(Inst, Seq)$ pairs from the fine-tuning phase. For a given test instruction, we calculate the \textbf{cosine similarity} between its semantic embedding and all instructions in $\mathcal{D}$. The top-3 most similar tasks are retrieved and provided to the VLM as \textbf{in-context priors}. By conditioning the VLM on the test instruction, the primitive library, the initial environment state (RGB), and these retrieved exemplars, we ensure that the generated plan remains grounded in the robot's known physical capabilities.

We also utilize \textbf{Qwen3-VL} as our base model. Here is an illustrative example of the prompt used to guide the VLM in Primitive Planner to generate execution flow with Retrieval-Augmented Planning strategy:

\begin{promptbox}[System Prompt: VLM-based Primitive Reasoning (Inference)] \\
\textbf{[Context]} \\
Given the \textbf{Initial Environment Image} and the \textbf{Full Task Instruction}, decompose the task into an ordered sequence of primitives and provide corresponding specific instructions for each.

\vspace{5pt}
\textbf{[Primitive Merging \& Selection Rules]}
\begin{itemize}[leftmargin=1.2em, label=$\bullet$, itemsep=0pt]
    \item \textbf{Motion Consolidation}: Movements independent of a specific object should be merged into the subsequent interaction. For example, use \textit{grasp} directly instead of \textit{move + grasp}.
    \item \textbf{Planar Motion Merging}: Two adjacent primitives primarily involving motion in the same plane (e.g., \textit{move}, \textit{lift}, \textit{push}) can be merged into a single segment.
    \item \textbf{Z-Axis Prioritization}: For tasks with significant height requirements (e.g., target position is elevated), prioritize using \textit{lift} or \textit{place} over horizontal primitives.
    \item \textbf{Multi-Object Labeling}: If the task involves multiple sub-goals (e.g., picking multiple objects), append numerical suffixes (\textit{\_1, \_2}) to repeated primitives to avoid ambiguity.
\end{itemize}

\vspace{5pt}
\textbf{[Primitive Definitions]}
\begin{itemize}[leftmargin=1.2em, label=$\diamond$, itemsep=0pt]
    \item \textbf{grasp}: Approaching, closing the gripper on an object, and slightly lifting.
    \item \textbf{lift}: Moving upward significantly in the $z$-direction while the gripper is closed to meet height constraints; may involve $xy$ displacement.
    \item \textbf{move}: Large-scale horizontal movement in the $xy$-plane while the gripper is closed.
    \item \textbf{place}: Descending to place an object at a target position followed by a departure lift.
    \item \textbf{push}: Pushing an object across a surface to a target location.
\end{itemize}

\vspace{5pt}
\textbf{[Retrieved Prior References (In-Context)]} \\
The following "Task $\rightarrow$ Primitive Sequence" mappings are retrieved from the Disassembly Library based on semantic similarity. Use them as references for typical decomposition strategies:
\begin{itemize}[leftmargin=1.2em, nosep]
    \item \textit{pick sponge and place on the plate}: [\texttt{grasp, move, place}]
    \item \textit{pick cup and place on the cabinet}: [\texttt{grasp, lift, place}]
    \item \textit{push in the bottom drawer}: [\texttt{push}]
\end{itemize}

\vspace{5pt}
\textbf{[Target Task]} \\
\textbf{Current Instruction}: "{task}" \\
\textbf{Initial Image}: [Visual input showing current scene configuration]

\vspace{5pt}
\textbf{[Expected Output]} \\
Based on the image and instruction, provide a reasonable primitive-level decomposition.
\textit{Output Format}: Reasoning $\rightarrow$ \texttt{[Prim\_1, Prim\_2, ...]}
\end{promptbox}

\subsection{Inference: Primitive Switch}
To ensure execution stability and minimize latency during inference, we employ a rule-based controller for primitive switching. Unlike the offline segmentation used during fine-tuning, the inference controller operates without access to future states. Consequently, the symmetric temporal window is replaced by a \textbf{history sliding window} that relies exclusively on historical proprioception and action data.

To maintain consistency between fine-tuning and inference, we provide the \textbf{original offline segmentation code} as a reference within the prompt. We utilize DeepSeek V3 to automatically simplify and adapt this logic into a lightweight, real-time switching code. This ensures that the physical transition triggers remain aligned across both phases. Here is an illustrative example of the prompt used to guide the LLM in generating \textbf{switch code} to control the execution flow planned by Primitive Planner:

\begin{promptbox}[System Prompt: Real-time Primitive Switching Generation]\\
\textbf{[System Role]} \\
You are a robotic software engineer. Your task is to generate a Python function \texttt{get\_next\_primitive} for online execution. You must \textbf{strictly adhere} to the original primitive definitions and segmentation conditions provided below, adapting them \textbf{only} in implementation to fit a 10-step history buffer.

\vspace{5pt}
\textbf{[Input Context]}
\begin{itemize}[leftmargin=1.5em, nosep]
    \item \textbf{History States} ($s_{t-9:t}$): Last 10 frames [7-dim: pos, axis-angle, gripper\_state].
    \item \textbf{History Actions} ($a_{t-9:t}$): Last 10 frames [7-dim: relative\_pos, relative\_axis-angle, gripper\_action].
    \item \textbf{Current Primitive}: The name of the primitive being executed.
    \item \textbf{Primitive Sequence}: The planned execution flow (list of strings).
\end{itemize}

\vspace{5pt}
\textbf{[Implementation Constraints]}
\begin{itemize}[leftmargin=1.2em, label=$\bullet$, itemsep=0pt]
    \item \textbf{Causal Logic Only}: You have NO access to future frames. The "Current Frame" in the offline logic corresponds to the \textbf{last frame} (index 9) of the input history.
    \item \textbf{Strict Condition Adherence}: Do not modify the physical thresholds (e.g., $0.02, 0.08, 0.005$) or the qualitative definitions from the reference code. 
    \item \textbf{Window Adaptation}: For conditions that originally required a future window, use the tail end of the 10-step history to verify if those trends have \textbf{already started to manifest} at the current moment $t$.
\end{itemize}

\vspace{5pt}
\textbf{[Reference Criteria]} \\
\textit{// You must use the exact logic for grasp, lift, move, place, and push as defined here: //}
\begin{enumerate}
\item \textbf{grasp}: Constant gripper state across the frames near the current frame and the "future" frames; gripper action remains in the "closed" state; significant horizontal or $z$-axis movement begins in the "future" frames.
    \item \textbf{lift}: Significant $z$-axis increase over several "past" frames in the buffer; the current/recent frames show a stop in rising, a continuous descending trend, or a cessation of movement in both $xy$ directions.
    \item \textbf{move}: Significant $x$ or $y$ movement over several "past" frames; movement in $xy$ directions tends to stop in recent frames, or the $z$-axis begins a sharp descent.
    \item \textbf{place}: Gripper state is "open" across recent and "past" frames, accompanied by a lift in the $z$ direction.
    \item \textbf{push}: Significant $x$ or $y$ movement over several "past" frames; $xy$ movement tends to stop in recent frames, with minimal change in the $z$ direction throughout.
\end{enumerate}

\vspace{5pt}
\textbf{[Reference Code]} \\
\textit{(Full Offline Code provided in context)}

\vspace{5pt}
\textbf{[Function Logic]}
\begin{itemize}[leftmargin=1.2em, label=$\rightarrow$]
    \item If \texttt{current\_primitive} is the last in \texttt{seq}: return \texttt{current\_primitive}.
    \item If \texttt{check\_[primitive]\_condition} returns \textit{True}: return \texttt{next\_primitive} from \texttt{seq}.
    \item Otherwise: return \texttt{current\_primitive}.
\end{itemize}

\vspace{5pt}
\textbf{[Output Format]} \\
Return the Python function: \texttt{def get\_next\_primitive(states, actions, curr\_name, seq):}
\end{promptbox}

\subsection{Multimodal Canonical Representation}
To harmonize diverse motion patterns across different tasks, we introduce the Multimodal Canonical Representation (MCR) as a standardized interface.

\textbf{1. Motivation for MCR}

Without MCR, primitive instructions must retain task-specific contextual information, such as specific actions, objects, and spatial relationships. This results in significant variations among instructions for the same primitive type across different tasks. Especially in scenarios involving multiple tasks within the same scene, the model is forced to over-focus on the idiosyncratic differences in instructions rather than the shared underlying motion pattern. This unnecessarily expands the instruction space—often making it larger than that of non-decomposed tasks—and hinders the model's ability to generalize to Out-of-Distribution (OOD) tasks, as shown in our ablation studies (w/o MCR (Disass. Only)).

MCR addresses this by reallocating task components: it unifies the instruction space to focus on the primitive type, while using object-centric masks as the primary carrier of task context. This ensures that the VLA inputs are representationally consistent, enabling cross-task experience sharing without losing the scene-specific information required for execution.

\textbf{2. Instruction Mapping}

When MCR is active, task-specific instructions are first mapped to standardized canonical ones based on their primitive type. The mapping table is as follows:

\begin{table}[ht]
\centering
\caption{\textbf{Complete MCR Primitive Instruction Mapping.} Note: \texttt{color\_1} and \texttt{color\_2} are custom color indices representing the semantic masks of the manipulated object and the target location/object, respectively.}
\label{tab:mcr_full_mapping}
\vspace{1em}
\begin{tabular}{lp{9cm}}
\hline
\textbf{Primitive Type} & \textbf{Canonical Instruction (MCR Interface)} \\ \hline
\rowcolor{gray!10} grasp & "Grasp the masked object with color\_1" \\
lift & "Lift the object with color\_1" \\
\rowcolor{gray!10} move & "Move to above the object with color\_2" \\
place & "Place in the object with color\_2" \\
\rowcolor{gray!10} push & "Push the object with color\_1" \\
press & "Press the object with color\_2" \\
\rowcolor{gray!10} pull & "Pull the object with color\_1" \\
insert & "Insert into the area with color\_1" \\
\rowcolor{gray!10} twist & "Twist the grasped object" \\
rotate & "Rotate the object with color\_1" \\
\rowcolor{gray!10} tilt & "Tilt the object with color\_1" \\ \hline
\end{tabular}
\end{table}

\textbf{3. Visual Grounding and Tracking}

For each task, we first extract the target objects mentioned in the original instruction. We utilize the \textbf{Qwen2.5-VL-72B-Instruct}~\cite{qwen2.5vl2025} model to perform zero-shot object \textbf{detection}. The model is queried with a specific prompt to provide the spatial coordinates of the objects:

\begin{promptbox}[System Prompt: Visual Object Detection]\\
\textbf{[System Role]} \\
You are an object detection agent. Given an image, your task is to identify and provide the coordinates of a specific target object.

\vspace{5pt}
\textbf{[Instruction]} \\
Find the bounding box of the following object: \texttt{"\{obj\}"}.

\vspace{5pt}
\textbf{[Input]} \\
\textbf{The image is}: [Visual input of the current environment]

\vspace{5pt}
\textbf{[Output Format]} \\
You must respond strictly in JSON format as follows:
\begin{minipage}{0.6\textwidth}
\begin{lstlisting}
{
  "bbox": [x1, y1, x2, y2]
}
\end{lstlisting}
\end{minipage}

\vspace{5pt}
\textit{Note: The coordinates [x1, y1, x2, y2] represent the top-left and bottom-right corners of the bounding box normalized to the image dimensions.}
\end{promptbox}

The VLM returns a JSON-formatted bounding box $[x_1, y_1, x_2, y_2]$, which identifies the initial location of the task-relevant object (assigned as \texttt{color\_1}) or the target area (assigned as \texttt{color\_2}).

Once the bounding box is obtained, the system proceeds with the following steps to ensure spatial consistency:
\begin{itemize}
\item \textbf{Segmentation}: The bounding box is passed to the \textbf{Segment Anything Model (SAM)}~\cite{kirillov2023segany}, which generates a high-fidelity initial mask for the object.
\item \textbf{Continuous Tracking}: To handle dynamic scenes and robot movements, the \textbf{Cutie}~\cite{cheng2024putting} video object segmentation model is employed. Cutie tracks the initial mask throughout the entire trajectory, providing real-time mask updates with minimal latency.
\end{itemize}

This multi-stage visual pipeline ensures both high-quality data generation and real-time execution efficiency through a decoupled deployment strategy.

For offline data processing, this automated workflow allows us to convert large-scale raw primitive trajectories into MCR-aligned trajectories with high semantic precision, eliminating the need for manual annotation.

For \textbf{online inference}, the computational overhead is strategically distributed: the heavy-duty Qwen2.5-VL-72B-Instruct and SAM are invoked \textbf{only once during a pre-execution initialization} phase to ground the task context. Once the motion begins, the system hands over the tracking responsibility to the \textbf{lightweight Cutie model}. This transition ensures that the MCR transformation maintains negligible latency during robot execution, allowing for responsive control and seamless alignment of task-specific observations even in unseen OOD scenarios (detailed in Section \ref{sec:latency_analysis}).

By combining these tracked masks with unified instructions, the MCR provides a standardized interface where the visual information within the mask supplies the necessary task context, ensuring the VLA can understand "where and what" to act upon despite the simplified text input.

\FloatBarrier

\section{Qualitative Comparison of Primitive Disassembly}
\label{appendix:Comparison of Disassembly}
\begin{figure}[tb]
    \centering
    \includegraphics[width=\columnwidth]{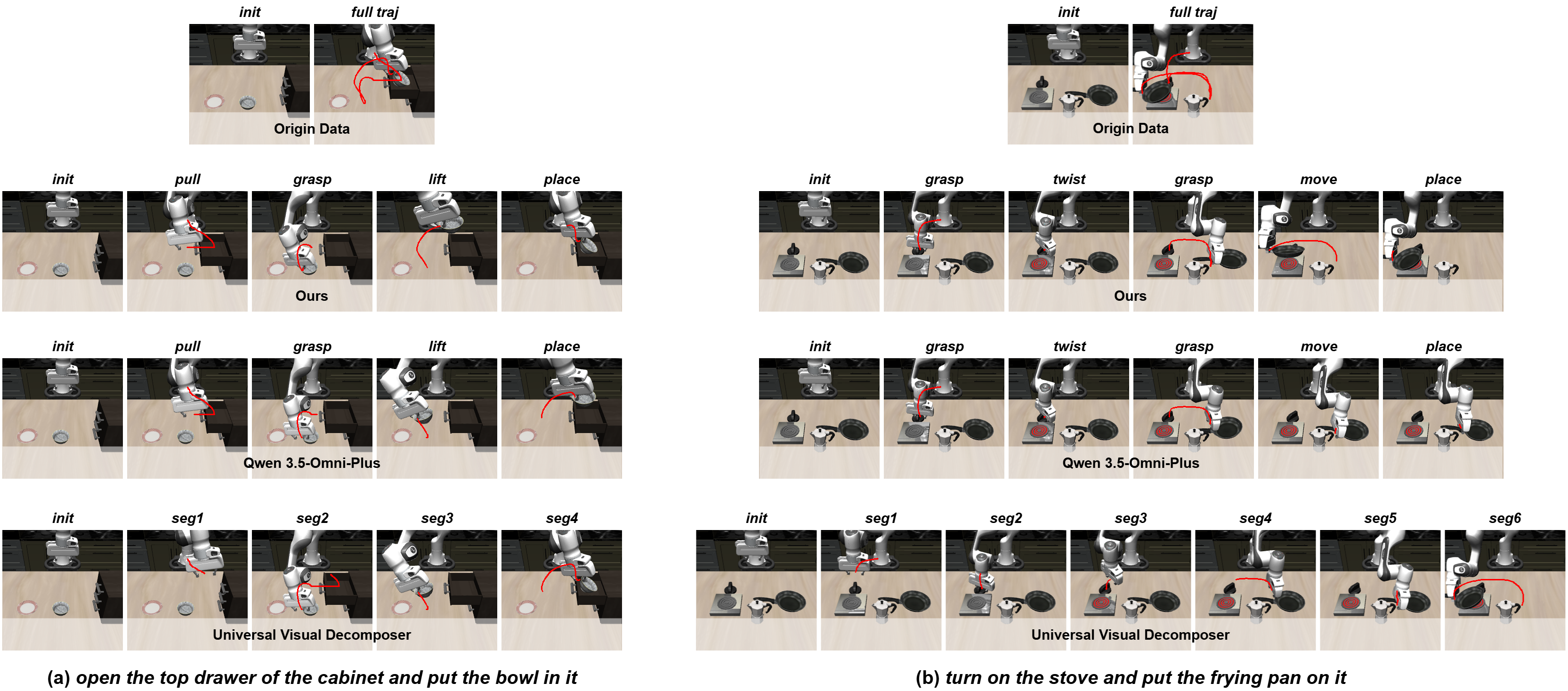}
    \caption{\textbf{Qualitative comparison of primitive disassembly methods on LIBERO-90.} We visualize the disassembly results on two complex tasks: (a) \textit{open the top drawer of the cabinet and put the bowl in it} and (b) \textit{turn on the stove and put the frying pan on it}. \quad
    Compared to the vision-only \textbf{Universal Visual Decomposer} which lacks semantic awareness (resulting in merged interactions in (a) and over-segmentation in (b)) and the \textbf{direct VLM-based method} (which exhibits imprecise boundaries in (a), mid-task termination and redundant "move" segments in (b)), our \textbf{PrimitiveVLA} achieves the most precise and stable disassembly. By combining VLM semantic reasoning with deterministic rule-based code, our framework extracts clean, reusable primitives that are semantically grounded and temporally accurate, providing a superior foundation for VLA policy training and hierarchical assembly.}
\label{fig:comparison}
\end{figure}

To further validate the efficacy of our method, we provide a qualitative comparison across two complex tasks from the LIBERO-90 benchmark: \textbf{(a) \textit{open the top drawer of the cabinet and put the bowl in it}} and \textbf{(b) \textit{turn on the stove and put the frying pan on it}}. As illustrated in Fig.~\ref{fig:comparison}, we evaluate our pipeline against the \textbf{Universal Visual Decomposer} (UVD)~\cite{zhang2024universal} and a state-of-the-art VLM, \textbf{Qwen3.5-Omni-Plus}~\cite{qwen_omni_2026}, focusing on the precision and reliability of their primitive disassembly results.

\textbf{Limitations of Vision-only Methods (UVD):}
As a self-supervised visual method lacking linguistic guidance, UVD fails to achieve semantic-aware disassembly. This approach frequently \textbf{misses key interaction boundaries}: in task \textbf{(a)}, UVD collapses "opening the drawer" and "grasping the bowl" into a single segment (seg2), failing to disassemble the trajectory into independent, reusable primitives. In task \textbf{(b)}, it over-segments a continuous movement into redundant parts (seg3 and seg4), while generating several \textbf{low-significance} fragments (e.g., seg1 in \textbf{(a)}; seg1 and seg5 in \textbf{(b)}). Such inconsistent outputs demonstrate that purely visual cues are insufficient for robust disassembly.

\textbf{Limitations of Direct VLM-based Disassembly (Qwen3.5-Omni-Plus):}
Even when provided with predefined primitive definitions, using a high-end VLM for disassembly faces three critical challenges:
\begin{enumerate}
\item \textbf{Imprecise Recognition:} The VLM exhibits noticeable flaws in \textbf{identifying the temporal boundaries} for disassembly, such as the "grasp" and "pull" transitions in task \textbf{(a)}. In task \textbf{(b)}, it fails to cover the complete task horizon and prematurely stops, leading to incomplete \textbf{disassembly}.
\item \textbf{Black-box Logic:} The VLM's criteria for disassembly remain opaque. This \textbf{lack of interpretability} means we cannot extract deterministic switching rules for subsequent inference. If the model suffers from hallucinations (as seen in the mid-task termination in \textbf{(b)}), the resulting \textbf{disassembly} will lead to execution failure.
\item \textbf{High Computational Cost:} Relying on proprietary VLMs to process large-scale datasets for disassembly is prohibitively \textbf{expensive and difficult to scale}.
\end{enumerate}

\textbf{Advantages of Our Method:}
In contrast, our approach leverages the VLM for high-level semantic disassembly while delegating the fine-grained temporal anchoring to \textbf{rule-based code}. This hybrid strategy ensures that all primitives are semantically grounded and \textbf{precisely} bounded, providing a \textbf{robust and interpretable} foundation for VLA training. By replacing expensive, opaque visual reasoning with deterministic code-based triggers, we achieve \textbf{high-precision, low-cost} disassembly that extracts reusable primitives with high stability across diverse robotic tasks.

\section{Experiment Setting Details}

\subsection{Detailed Task Lists}
\label{appendix:task_lists}

\vspace{-2pt}
\begin{figure}[tbh]
    \centering
    \includegraphics[width=\columnwidth]{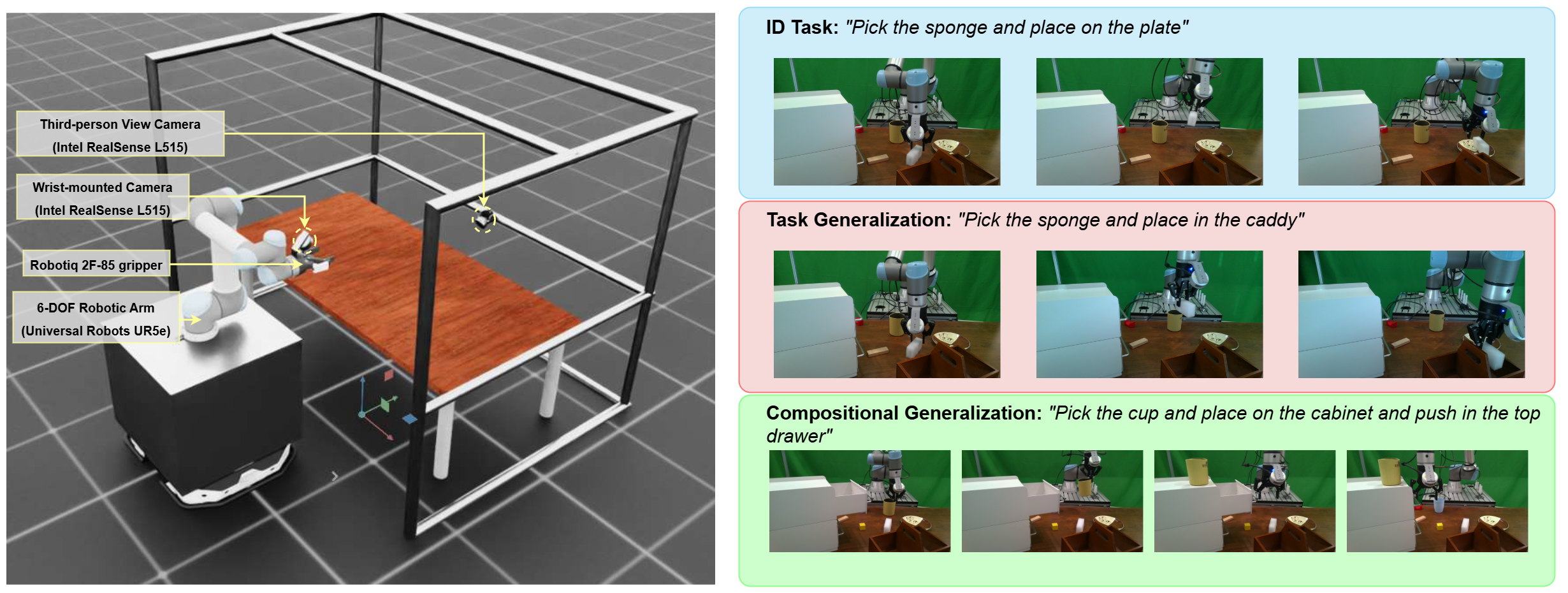}
    \caption{\textbf{Real-World Experimental Setup and Evaluation Sequences.} \textbf{Left:} \textbf{The physical workspace setup}, demonstrating the integrated arrangement of the robotic arm, the multi-view camera system (third-person and wrist-mounted), and the operational tabletop layout. \textbf{Right:} Representative rollout sequences illustrating successful executions across three distinct evaluation categories: standard In-Distribution (ID) tasks, novel Task Generalization scenarios, and complex Compositional Generalization tasks.}
    \label{fig:real_world_setup}
    \vspace{-5pt}
\end{figure}

\noindent
The tasks utilized in our experiments span various simulation environments and real-world scenarios (summarized in Table~\ref{tab:full_task_table}). To provide a clear mapping between high-level instructions and executable actions, we also include the disassembly results (i.e., the corresponding primitive sequences) for each task in the same table. The task design logic is as follows:

\begin{itemize}
    \item \textbf{In-Distribution and Large-scale fine-tuning Benchmarks (Simulation ID):} This category includes \texttt{LIBERO-Spatial}, \texttt{LIBERO-Object}, and \texttt{LIBERO-Goal}~\cite{liu2023libero}. Among them, \texttt{LIBERO-90} (All tasks listed in Table~\ref{tab:libero90_subset}) serves as the primary benchmark for large-scale fine-tuning, used to train the model's fundamental action representations and policy learning.
    \item \textbf{Complex Distribution Adaptation Evaluation (RLBench):} Consistent with the fine-tuning evaluation protocol of LIBERO, RLBench~\cite{james2020rlbench} also adopts a "fine-tune then test" paradigm. Given that the object distribution in LIBERO is relatively consistent, we introduce RLBench to examine the model's ability to learn and execute tasks in environments with higher complexity, more diverse distributions, and multifaceted interaction types (e.g., involving higher degrees of freedom).
    \item \textbf{Zero-shot Evaluation Benchmarks (Zero-shot OOD):} \texttt{LIBERO-90 Novel} and \texttt{LIBERO-Long} are specifically employed to evaluate the zero-shot generalization capabilities of the model after fine-tuning on \texttt{LIBERO-90}. To ensure a fair and rigorous assessment of out-of-distribution (OOD) performance, the evaluation configurations—including the prompt templates, the Disassembly Library, and the control primitive switching code—remain strictly consistent with the settings used for the \texttt{LIBERO-90} experiments.
    \begin{itemize}
        \item \textbf{\texttt{LIBERO-90-Novel}:} Designed by recombining scenes and objects from LIBERO-90, this suite challenges the model to perform tasks on novel objects, execute long-horizon tasks formed by stitching two sub-tasks through a common component, and perform existing tasks under varying environmental configurations.
        \item \textbf{\texttt{LIBERO-Long} (LIBERO-10):} Serving as an OOD long-horizon benchmark, this suite assesses the model's zero-shot compositional reasoning for complex, multi-stage task sequences. In our evaluation, we select 8 tasks from this suite (10 tasks) that align with the manipulation learned from the \texttt{LIBERO-90} training set; the remaining two tasks are excluded as they require novel manipulation.
    \end{itemize}
    \item \textbf{Real-world Experiments (Real-world Exp):} Conducted in a laboratory setting to validate performance in physical scenarios. As illustrated in Fig.~\ref{fig:real_world_setup}, we present the physical workspace setup—comprising the robotic arm, the multi-view camera system (third-person and wrist-mounted), and the operational tabletop layout—alongside example rollouts of specific tasks. This includes 6 \textit{ID tasks} for data collection and fine-tuning, and 5 \textit{OOD tasks} designed to test zero-shot generalization (e.g., unseen object pairings) and long-horizon logic composition. To maintain experimental integrity, all configurations for the OOD tasks—including prompt templates, the Disassembly Library, and control primitive logic—are kept strictly consistent with those employed in the ID tasks.
\end{itemize}

\newcolumntype{L}{>{\raggedright\arraybackslash}X}

\begin{table}[htbp]
\centering
\caption{Comprehensive Task List for All Benchmarks.}
\label{tab:full_task_table}
\vspace{1em}
\renewcommand{\arraystretch}{0.75} %
\begin{tabularx}{\textwidth}{l >{\hsize=1.5\hsize}L >{\hsize=0.5\hsize}L}
\toprule
\textbf{Suite} & \textbf{Task Instruction (Full Name)} & \textbf{Primitive Sequence} \\ 
\midrule

\textbf{Libero-Spatial} 
& \fontsize{6}{7}\selectfont pick up the black bowl between the plate and the ramekin and place it on the plate & \fontsize{6}{7}\selectfont [grasp, move, place] \\[-0.5ex]
& \fontsize{6}{7}\selectfont pick up the black bowl from table center and place it on the plate & \fontsize{6}{7}\selectfont [grasp, move, place] \\[-0.5ex]
& \fontsize{6}{7}\selectfont pick up the black bowl in the top drawer of the wooden cabinet and place it on the plate & \fontsize{6}{7}\selectfont [grasp, move, place] \\[-0.5ex]
& \fontsize{6}{7}\selectfont pick up the black bowl next to the cookie box and place it on the plate & \fontsize{6}{7}\selectfont [grasp, move, place] \\[-0.5ex]
& \fontsize{6}{7}\selectfont pick up the black bowl next to the plate and place it on the plate & \fontsize{6}{7}\selectfont [grasp, move, place] \\[-0.5ex]
& \fontsize{6}{7}\selectfont pick up the black bowl next to the ramekin and place it on the plate & \fontsize{6}{7}\selectfont [grasp, move, place] \\[-0.5ex]
& \fontsize{6}{7}\selectfont pick up the black bowl on the cookie box and place it on the plate & \fontsize{6}{7}\selectfont [grasp, move, place] \\[-0.5ex]
& \fontsize{6}{7}\selectfont pick up the black bowl on the ramekin and place it on the plate & \fontsize{6}{7}\selectfont [grasp, move, place] \\[-0.5ex]
& \fontsize{6}{7}\selectfont pick up the black bowl on the stove and place it on the plate & \fontsize{6}{7}\selectfont [grasp, move, place] \\[-0.5ex]
& \fontsize{6}{7}\selectfont pick up the black bowl on the wooden cabinet and place it on the plate & \fontsize{6}{7}\selectfont [grasp, move, place] \\ \midrule

\textbf{Libero-Object} 
& \fontsize{6}{7}\selectfont pick up the alphabet soup and place it in the basket & \fontsize{6}{7}\selectfont [grasp, move, place] \\[-0.5ex]
& \fontsize{6}{7}\selectfont pick up the bbq sauce and place it in the basket & \fontsize{6}{7}\selectfont [grasp, move, place] \\[-0.5ex]
& \fontsize{6}{7}\selectfont pick up the butter and place it in the basket & \fontsize{6}{7}\selectfont [grasp, move, place] \\[-0.5ex]
& \fontsize{6}{7}\selectfont pick up the chocolate pudding and place it in the basket & \fontsize{6}{7}\selectfont [grasp, move, place] \\[-0.5ex]
& \fontsize{6}{7}\selectfont pick up the cream cheese and place it in the basket & \fontsize{6}{7}\selectfont [grasp, move, place] \\[-0.5ex]
& \fontsize{6}{7}\selectfont pick up the ketchup and place it in the basket & \fontsize{6}{7}\selectfont [grasp, move, place] \\[-0.5ex]
& \fontsize{6}{7}\selectfont pick up the milk and place it in the basket & \fontsize{6}{7}\selectfont [grasp, move, place] \\[-0.5ex]
& \fontsize{6}{7}\selectfont pick up the orange juice and place it in the basket & \fontsize{6}{7}\selectfont [grasp, move, place] \\[-0.5ex]
& \fontsize{6}{7}\selectfont pick up the salad dressing and place it in the basket & \fontsize{6}{7}\selectfont [grasp, move, place] \\[-0.5ex]
& \fontsize{6}{7}\selectfont pick up the tomato sauce and place it in the basket & \fontsize{6}{7}\selectfont [grasp, move, place] \\ \midrule

\textbf{Libero-Goal} 
& \fontsize{6}{7}\selectfont open the middle drawer of the cabinet & \fontsize{6}{7}\selectfont [pull] \\[-0.5ex]
& \fontsize{6}{7}\selectfont open the top drawer and put the bowl inside & \fontsize{6}{7}\selectfont [pull, grasp, lift, place] \\[-0.5ex]
& \fontsize{6}{7}\selectfont push the plate to the front of the stove & \fontsize{6}{7}\selectfont [press, push] \\[-0.5ex]
& \fontsize{6}{7}\selectfont put the bowl on the plate & \fontsize{6}{7}\selectfont [grasp, move, place] \\[-0.5ex]
& \fontsize{6}{7}\selectfont put the bowl on the stove & \fontsize{6}{7}\selectfont [grasp, move, place] \\[-0.5ex]
& \fontsize{6}{7}\selectfont put the bowl on top of the cabinet & \fontsize{6}{7}\selectfont [grasp, lift, place] \\[-0.5ex]
& \fontsize{6}{7}\selectfont put the cream cheese in the bowl & \fontsize{6}{7}\selectfont [grasp, move, place] \\[-0.5ex]
& \fontsize{6}{7}\selectfont put the wine bottle on the rack & \fontsize{6}{7}\selectfont [grasp, lift, place] \\[-0.5ex]
& \fontsize{6}{7}\selectfont put the wine bottle on top of the cabinet & \fontsize{6}{7}\selectfont [grasp, lift, place] \\[-0.5ex]
& \fontsize{6}{7}\selectfont turn on the stove & \fontsize{6}{7}\selectfont [grasp, twist] \\ \midrule

\textbf{Libero-90} 
& \fontsize{6}{7}\selectfont LIVING\_ROOM\_SCENE1\_put\_the\_black\_bowl\_on\_the\_plate & \fontsize{6}{7}\selectfont [grasp, move, place] \\[-0.5ex]
& \fontsize{6}{7}\selectfont LIVING\_ROOM\_SCENE4\_stack\_left\_bowl\_on\_right\_bowl\_and\_place\_them\_in\allowbreak\_tray & \fontsize{6}{7}\selectfont [grasp, move, place, grasp, move, place] \\[-0.5ex]
& \fontsize{6}{7}\selectfont KITCHEN\_SCENE3\_turn\_on\_the\_stove\_and\_put\_the\_frying\_pan\_on\_it & \fontsize{6}{7}\selectfont [grasp, twist, grasp, move, place] \\[-0.5ex]
& \fontsize{6}{7}\selectfont KITCHEN\_SCENE10\_put\_the\_butter\_in\_the\_top\_drawer\_and\_close\_it & \fontsize{6}{7}\selectfont [grasp, lift, place, push] \\[-0.5ex]
& \fontsize{6}{7}\selectfont STUDY\_SCENE4\_put\_the\_middle\_book\_on\_the\_shelf & \fontsize{6}{7}\selectfont [grasp, lift, tilt] \\[-0.5ex]
& \fontsize{6}{7}\selectfont ... (Total 84 tasks in this suite) & \fontsize{6}{7}\selectfont \vdots \\ \midrule

\textbf{Libero-90-Novel} 
& \fontsize{6}{7}\selectfont KITCHEN\_SCENE1\_open\_the\_top\_drawer\_put\_the\_black\_bowl\_in\_it\_and\_close\_it & \fontsize{6}{7}\selectfont [pull, grasp, lift, place, push] \\[-0.5ex]
& \fontsize{6}{7}\selectfont KITCHEN\_SCENE2\_stack\_the\_front\_bowl\_on\_the\_back\_bowl & \fontsize{6}{7}\selectfont [grasp, move, place] \\[-0.5ex]
& \fontsize{6}{7}\selectfont KITCHEN\_SCENE10\_open\_the\_top\_drawer\_put\_the\_chocolate\_pudding\_in\_it\_and\_close\_it & \fontsize{6}{7}\selectfont [pull, grasp, lift, place, push] \\[-0.5ex]
& \fontsize{6}{7}\selectfont KITCHEN\_SCENE10\_close\_the\_middle\_drawer & \fontsize{6}{7}\selectfont [push] \\[-0.5ex]
& \fontsize{6}{7}\selectfont KITCHEN\_SCENE10\_put\_the\_black\_bowl\_on\_the\_wooden\_cabinet & \fontsize{6}{7}\selectfont [grasp, lift, place] \\[-0.5ex]
& \fontsize{6}{7}\selectfont LIVING\_ROOM\_SCENE4\_pick\_up\_the\_right\_bowl\_and\_place\_it\_in\_the\_tray & \fontsize{6}{7}\selectfont [grasp, move, place] \\[-0.5ex]
& \fontsize{6}{7}\selectfont LIVING\_ROOM\_SCENE5\_put\_the\_yellow\_and\_white\_mug\_on\_the\_left\_plate & \fontsize{6}{7}\selectfont [grasp, move, place] \\[-0.5ex]
& \fontsize{6}{7}\selectfont LIVING\_ROOM\_SCENE2\_pick\_up\_the\_butter\_and\_place\_it\_in\_the\_basket & \fontsize{6}{7}\selectfont [grasp, move, place] \\ \midrule

\textbf{Libero-Long} 
& \fontsize{6}{7}\selectfont turn on the stove and put the moka pot on it & \fontsize{6}{7}\selectfont [grasp, twist, grasp, move, place] \\[-0.5ex]
& \fontsize{6}{7}\selectfont put the black bowl in the bottom drawer of the cabinet and close it & \fontsize{6}{7}\selectfont [grasp, move, place, push] \\[-0.5ex]
& \fontsize{6}{7}\selectfont put both moka pots on the stove & \fontsize{6}{7}\selectfont [grasp, move, place, grasp, move, place] \\[-0.5ex]
& \fontsize{6}{7}\selectfont put both the alphabet soup and the cream cheese box in the basket & \fontsize{6}{7}\selectfont [grasp, move, place, grasp, move, place] \\[-0.5ex]
& \fontsize{6}{7}\selectfont put both the alphabet soup and the tomato sauce in the basket & \fontsize{6}{7}\selectfont [grasp, move, place, grasp, move, place] \\[-0.5ex]
& \fontsize{6}{7}\selectfont put both the cream cheese box and the butter in the basket & \fontsize{6}{7}\selectfont [grasp, move, place, grasp, move, place] \\[-0.5ex]
& \fontsize{6}{7}\selectfont put the white mug on the left plate and put the yellow and white mug on the right plate & \fontsize{6}{7}\selectfont [grasp, move, place, grasp, move, place] \\[-0.5ex]
& \fontsize{6}{7}\selectfont put the white mug on the plate and put the chocolate pudding to the right of the plate & \fontsize{6}{7}\selectfont [grasp, move, place, grasp, move, place] \\ \midrule

\textbf{RLBench} 
& \fontsize{6}{7}\selectfont take\_umbrella\_out\_of\_umbrella\_stand & \fontsize{6}{7}\selectfont [grasp, lift] \\[-0.5ex]
& \fontsize{6}{7}\selectfont open\_jar & \fontsize{6}{7}\selectfont [grasp, twist, move, place] \\[-0.5ex]
& \fontsize{6}{7}\selectfont lamp\_off & \fontsize{6}{7}\selectfont [press] \\[-0.5ex]
& \fontsize{6}{7}\selectfont lift\_numbered\_block & \fontsize{6}{7}\selectfont [grasp, lift] \\[-0.5ex]
& \fontsize{6}{7}\selectfont open\_wine\_bottle & \fontsize{6}{7}\selectfont [grasp, twist, lift] \\[-0.5ex]
& \fontsize{6}{7}\selectfont push\_button & \fontsize{6}{7}\selectfont [press] \\[-0.5ex]
& \fontsize{6}{7}\selectfont put\_rubbish\_in\_bin & \fontsize{6}{7}\selectfont [grasp, lift, place] \\[-0.5ex]
& \fontsize{6}{7}\selectfont meat\_off\_grill & \fontsize{6}{7}\selectfont [grasp, move, place] \\[-0.5ex]
& \fontsize{6}{7}\selectfont take\_off\_weighing\_scales & \fontsize{6}{7}\selectfont [grasp, lift, move, place] \\[-0.5ex]
& \fontsize{6}{7}\selectfont take\_lid\_off\_saucepan & \fontsize{6}{7}\selectfont [grasp, lift] \\ \midrule

\multirow{13}{*}{\textbf{Real-World}} 
& \fontsize{6}{7}\selectfont \textbf{In-Distribution (ID):} & \\[-0.6ex]
& \fontsize{6}{7}\selectfont Pick sponge and place on the plate & \fontsize{6}{7}\selectfont [grasp, move, place] \\[-0.6ex]
& \fontsize{6}{7}\selectfont Pick cup and place on the top of cabinet & \fontsize{6}{7}\selectfont [grasp, lift, place] \\[-0.6ex]
& \fontsize{6}{7}\selectfont Pick cube and place in the top drawer & \fontsize{6}{7}\selectfont [grasp, lift, place] \\[-0.6ex]
& \fontsize{6}{7}\selectfont Pick cube and place in the left caddy & \fontsize{6}{7}\selectfont [grasp, move, place] \\[-0.6ex]
& \fontsize{6}{7}\selectfont Push in the top drawer & \fontsize{6}{7}\selectfont [push] \\[-0.6ex]
& \fontsize{6}{7}\selectfont Push in the bottom drawer & \fontsize{6}{7}\selectfont [push] \\ \cmidrule{2-3}

& \fontsize{6}{7}\selectfont \textbf{Generalization (OOD):} & \\[-0.6ex]
& \fontsize{6}{7}\selectfont Pick sponge and place in the left caddy (unseen pair) & \fontsize{6}{7}\selectfont [grasp, move, place] \\[-0.6ex]
& \fontsize{6}{7}\selectfont Pick cube and place on the top of cabinet (unseen pair) & \fontsize{6}{7}\selectfont [grasp, lift, place] \\[-0.6ex]
& \fontsize{6}{7}\selectfont Pick the blue cup and place on the top of cabinet (novel object) & \fontsize{6}{7}\selectfont [grasp, lift, place] \\ \cmidrule{2-3}

& \fontsize{6}{7}\selectfont \textbf{Compositional (OOD):} & \\[-0.6ex]
& \fontsize{6}{7}\selectfont Pick cup then push drawer & \fontsize{6}{7}\selectfont [grasp, lift, place, push] \\[-0.6ex]
& \fontsize{6}{7}\selectfont Pick sponge and place on the plate, then Pick cube and place in the left caddy & \fontsize{6}{7}\selectfont [grasp, move, place, grasp, move, place] \\ 

\bottomrule
\end{tabularx}
\end{table}

\begin{table}[htbp]
\centering
\caption{Detailed List of the Libero-90 Subset with Primitive Sequences.}
\label{tab:libero90_subset}
\vspace{1em}
\fontsize{6}{7}\selectfont %
\renewcommand{\arraystretch}{0.8}
\begin{tabularx}{\textwidth}{l >{\hsize=1.45\hsize}L >{\hsize=0.55\hsize}L}
\toprule
\textbf{ID} & \textbf{Task Instruction (Full Name)} & \textbf{Primitive Sequence} \\ 
\midrule
1 & LIVING\_ROOM\_SCENE1\_pick\_up\_the\_alphabet\_soup\_and\_put\_it\_in\_the\_basket & [grasp, move, place] \\
2 & LIVING\_ROOM\_SCENE2\_pick\_up\_the\_alphabet\_soup\_and\_put\_it\_in\_the\_basket & [grasp, move, place] \\
3 & LIVING\_ROOM\_SCENE3\_pick\_up\_the\_alphabet\_soup\_and\_put\_it\_in\_the\_tray & [grasp, move, place] \\
4 & LIVING\_ROOM\_SCENE4\_pick\_up\_the\_chocolate\_pudding\_and\_put\_it\_in\_the\_tray & [grasp, move, place] \\
5 & LIVING\_ROOM\_SCENE1\_pick\_up\_the\_cream\_cheese\_box\_and\_put\_it\_in\_the\_basket & [grasp, move, place] \\
6 & LIVING\_ROOM\_SCENE3\_pick\_up\_the\_cream\_cheese\_and\_put\_it\_in\_the\_tray & [grasp, move, place] \\
7 & LIVING\_ROOM\_SCENE1\_pick\_up\_the\_ketchup\_and\_put\_it\_in\_the\_basket & [grasp, move, place] \\
8 & LIVING\_ROOM\_SCENE3\_pick\_up\_the\_ketchup\_and\_put\_it\_in\_the\_tray & [grasp, move, place] \\
9 & LIVING\_ROOM\_SCENE2\_pick\_up\_the\_butter\_and\_put\_it\_in\_the\_basket & [grasp, move, place] \\
10 & LIVING\_ROOM\_SCENE3\_pick\_up\_the\_butter\_and\_put\_it\_in\_the\_tray & [grasp, move, place] \\
11 & LIVING\_ROOM\_SCENE2\_pick\_up\_the\_milk\_and\_put\_it\_in\_the\_basket & [grasp, move, place] \\
12 & LIVING\_ROOM\_SCENE2\_pick\_up\_the\_orange\_juice\_and\_put\_it\_in\_the\_basket & [grasp, move, place] \\
13 & LIVING\_ROOM\_SCENE4\_pick\_up\_the\_salad\_dressing\_and\_put\_it\_in\_the\_tray & [grasp, move, place] \\
14 & LIVING\_ROOM\_SCENE1\_pick\_up\_the\_tomato\_sauce\_and\_put\_it\_in\_the\_basket & [grasp, move, place] \\
15 & LIVING\_ROOM\_SCENE2\_pick\_up\_the\_tomato\_sauce\_and\_put\_it\_in\_the\_basket & [grasp, move, place] \\
16 & LIVING\_ROOM\_SCENE3\_pick\_up\_the\_tomato\_sauce\_and\_put\_it\_in\_the\_tray & [grasp, move, place] \\
17 & LIVING\_ROOM\_SCENE4\_pick\_up\_the\_black\_bowl\_on\_the\_left\_and\_put\_it\_in\_the\_tray & [grasp, move, place] \\
18 & LIVING\_ROOM\_SCENE4\_stack\_the\_left\_bowl\_on\_the\_right\_bowl\_and\_place\_them\_in\_the\_tray & [grasp, move, place, grasp, move, place] \\
19 & LIVING\_ROOM\_SCENE4\_stack\_the\_right\_bowl\_on\_the\_left\_bowl\_and\_place\_them\_in\_the\_tray & [grasp, move, place, grasp, move, place] \\
20 & LIVING\_ROOM\_SCENE5\_put\_the\_red\_mug\_on\_the\_left\_plate & [grasp, move, place] \\
21 & LIVING\_ROOM\_SCENE5\_put\_the\_red\_mug\_on\_the\_right\_plate & [grasp, move, place] \\
22 & LIVING\_ROOM\_SCENE5\_put\_the\_white\_mug\_on\_the\_left\_plate & [grasp, move, place] \\
23 & LIVING\_ROOM\_SCENE5\_put\_the\_yellow\_and\_white\_mug\_on\_the\_right\_plate & [grasp, move, place] \\
24 & LIVING\_ROOM\_SCENE6\_put\_the\_chocolate\_pudding\_to\_the\_right\_of\_the\_plate & [grasp, move, place] \\
25 & LIVING\_ROOM\_SCENE6\_put\_the\_red\_mug\_on\_the\_plate & [grasp, move, place] \\
26 & LIVING\_ROOM\_SCENE6\_put\_the\_white\_mug\_on\_the\_plate & [grasp, move, place] \\
27 & KITCHEN\_SCENE1\_put\_the\_black\_bowl\_on\_the\_plate & [grasp, move, place] \\
28 & KITCHEN\_SCENE3\_put\_the\_frying\_pan\_on\_the\_stove & [grasp, move, place] \\
29 & KITCHEN\_SCENE3\_put\_the\_moka\_pot\_on\_the\_stove & [grasp, move, place] \\
30 & KITCHEN\_SCENE1\_put\_the\_black\_bowl\_on\_top\_of\_the\_cabinet & [grasp, lift, place] \\
31 & KITCHEN\_SCENE2\_put\_the\_black\_bowl\_at\_the\_back\_on\_the\_plate & [grasp, move, place] \\
32 & KITCHEN\_SCENE2\_put\_the\_black\_bowl\_at\_the\_front\_on\_the\_plate & [grasp, move, place] \\
33 & KITCHEN\_SCENE2\_put\_the\_middle\_black\_bowl\_on\_the\_plate & [grasp, move, place] \\
34 & KITCHEN\_SCENE2\_put\_the\_middle\_black\_bowl\_on\_top\_of\_the\_cabinet & [grasp, lift, place] \\
35 & KITCHEN\_SCENE2\_stack\_the\_black\_bowl\_at\_the\_front\_on\_the\_black\_bowl\_in\_the\_middle & [grasp, move, place] \\
36 & KITCHEN\_SCENE2\_stack\_the\_middle\_black\_bowl\_on\_the\_back\_black\_bowl & [grasp, move, place] \\
37 & KITCHEN\_SCENE4\_put\_the\_black\_bowl\_in\_the\_bottom\_drawer\_of\_the\_cabinet & [grasp, move, place] \\
38 & KITCHEN\_SCENE4\_put\_the\_black\_bowl\_on\_top\_of\_the\_cabinet & [grasp, lift, place] \\
39 & KITCHEN\_SCENE4\_put\_the\_wine\_bottle\_on\_the\_wine\_rack & [grasp, lift, place] \\
40 & KITCHEN\_SCENE3\_turn\_on\_the\_stove & [grasp, twist] \\
41 & KITCHEN\_SCENE3\_turn\_on\_the\_stove\_and\_put\_the\_frying\_pan\_on\_it & [grasp, twist, grasp, move, place] \\
42 & KITCHEN\_SCENE1\_open\_the\_top\_drawer\_of\_the\_cabinet\_and\_put\_the\_bowl\_in\_it & [pull, grasp, lift, place] \\
43 & KITCHEN\_SCENE5\_put\_the\_black\_bowl\_in\_the\_top\_drawer\_of\_the\_cabinet & [grasp, lift, place] \\
44 & KITCHEN\_SCENE5\_put\_the\_black\_bowl\_on\_the\_plate & [grasp, move, place] \\
45 & KITCHEN\_SCENE5\_put\_the\_black\_bowl\_on\_top\_of\_the\_cabinet & [grasp, lift, place] \\
46 & KITCHEN\_SCENE7\_put\_the\_white\_bowl\_on\_the\_plate & [grasp, move, place] \\
47 & KITCHEN\_SCENE7\_put\_the\_white\_bowl\_to\_the\_right\_of\_the\_plate & [grasp, move, place] \\
48 & KITCHEN\_SCENE8\_put\_the\_right\_moka\_pot\_on\_the\_stove & [grasp, move, place] \\
49 & KITCHEN\_SCENE8\_turn\_off\_the\_stove & [grasp, twist] \\
50 & KITCHEN\_SCENE9\_put\_the\_frying\_pan\_on\_the\_cabinet\_shelf & [grasp, lift, insert] \\
51 & KITCHEN\_SCENE9\_put\_the\_frying\_pan\_on\_top\_of\_the\_cabinet & [grasp, lift, place] \\
52 & KITCHEN\_SCENE9\_put\_the\_frying\_pan\_under\_the\_cabinet\_shelf & [grasp, insert] \\
53 & KITCHEN\_SCENE9\_put\_the\_white\_bowl\_on\_top\_of\_the\_cabinet & [grasp, lift, place] \\
54 & KITCHEN\_SCENE9\_turn\_on\_the\_stove & [grasp, twist] \\
55 & KITCHEN\_SCENE9\_turn\_on\_the\_stove\_and\_put\_the\_frying\_pan\_on\_it & [grasp, twist, grasp, move, place] \\
56 & KITCHEN\_SCENE10\_close\_the\_top\_drawer\_of\_the\_cabinet\_and\_put\_the\_black\_bowl\_on\_top\_of\_it & [push, grasp, lift, place] \\
57 & KITCHEN\_SCENE10\_put\_the\_black\_bowl\_in\_the\_top\_drawer\_of\_the\_cabinet & [grasp, lift, place] \\
58 & KITCHEN\_SCENE10\_put\_the\_butter\_at\_the\_back\_in\_the\_top\_drawer\_of\_the\_cabinet\_and\_close\_it & [grasp, lift, place, push] \\
59 & KITCHEN\_SCENE10\_put\_the\_butter\_at\_the\_front\_in\_the\_top\_drawer\_of\_the\_cabinet\_and\_close\_it & [grasp, lift, place, push] \\
60 & KITCHEN\_SCENE10\_put\_the\_chocolate\_pudding\_in\_the\_top\_drawer\_of\_the\_cabinet\_and\_close\_it & [grasp, lift, place, push] \\
61 & STUDY\_SCENE1\_pick\_up\_the\_book\_and\_place\_it\_in\_the\_front\_compartment\_of\_the\_caddy & [grasp, lift, insert] \\
62 & STUDY\_SCENE1\_pick\_up\_the\_book\_and\_place\_it\_in\_the\_left\_compartment\_of\_the\_caddy & [grasp, lift, insert] \\
63 & STUDY\_SCENE1\_pick\_up\_the\_book\_and\_place\_it\_in\_the\_right\_compartment\_of\_the\_caddy & [grasp, lift, insert] \\
64 & STUDY\_SCENE2\_pick\_up\_the\_book\_and\_place\_it\_in\_the\_back\_compartment\_of\_the\_caddy & [grasp, lift, insert] \\
65 & STUDY\_SCENE2\_pick\_up\_the\_book\_and\_place\_it\_in\_the\_front\_compartment\_of\_the\_caddy & [grasp, lift, insert] \\
66 & STUDY\_SCENE2\_pick\_up\_the\_book\_and\_place\_it\_in\_the\_left\_compartment\_of\_the\_caddy & [grasp, lift, insert] \\
67 & STUDY\_SCENE2\_pick\_up\_the\_book\_and\_place\_it\_in\_the\_right\_compartment\_of\_the\_caddy & [grasp, lift, insert] \\
68 & STUDY\_SCENE3\_pick\_up\_the\_book\_and\_place\_it\_in\_the\_front\_compartment\_of\_the\_caddy & [grasp, lift, insert] \\
69 & STUDY\_SCENE3\_pick\_up\_the\_book\_and\_place\_it\_in\_the\_left\_compartment\_of\_the\_caddy & [grasp, lift, insert] \\
70 & STUDY\_SCENE3\_pick\_up\_the\_book\_and\_place\_it\_in\_the\_right\_compartment\_of\_the\_caddy & [grasp, lift, insert] \\
71 & STUDY\_SCENE1\_pick\_up\_the\_yellow\_and\_white\_mug\_and\_place\_it\_to\_the\_right\_of\_the\_caddy & [grasp, move, place] \\
72 & STUDY\_SCENE3\_pick\_up\_the\_red\_mug\_and\_place\_it\_to\_the\_right\_of\_the\_caddy & [grasp, move, place] \\
73 & STUDY\_SCENE3\_pick\_up\_the\_white\_mug\_and\_place\_it\_to\_the\_right\_of\_the\_caddy & [grasp, move, place] \\
74 & STUDY\_SCENE4\_pick\_up\_the\_book\_in\_the\_middle\_and\_place\_it\_on\_the\_cabinet\_shelf & [grasp, lift, tilt] \\
75 & STUDY\_SCENE4\_pick\_up\_the\_book\_on\_the\_left\_and\_place\_it\_on\_top\_of\_the\_shelf & [grasp, lift, tilt] \\
76 & STUDY\_SCENE4\_pick\_up\_the\_book\_on\_the\_right\_and\_place\_it\_on\_the\_cabinet\_shelf & [grasp, tilt] \\
77 & STUDY\_SCENE4\_pick\_up\_the\_book\_on\_the\_right\_and\_place\_it\_under\_the\_cabinet\_shelf & [grasp, tilt] \\
78 & KITCHEN\_SCENE1\_open\_the\_bottom\_drawer\_of\_the\_cabinet & [pull] \\
79 & KITCHEN\_SCENE1\_open\_the\_top\_drawer\_of\_the\_cabinet & [pull] \\
80 & KITCHEN\_SCENE2\_open\_the\_top\_drawer\_of\_the\_cabinet & [pull] \\
81 & KITCHEN\_SCENE4\_close\_the\_bottom\_drawer\_of\_the\_cabinet & [push] \\
82 & KITCHEN\_SCENE4\_close\_the\_bottom\_drawer\_of\_the\_cabinet\_and\_open\_the\_top\_drawer & [push, pull] \\
83 & KITCHEN\_SCENE5\_close\_the\_top\_drawer\_of\_the\_cabinet & [push] \\
84 & KITCHEN\_SCENE10\_close\_the\_top\_drawer\_of\_the\_cabinet & [push] \\
85 & KITCHEN\_SCENE7\_open\_the\_microwave & [rotate] \\
86 & KITCHEN\_SCENE5\_put\_the\_ketchup\_in\_the\_top\_drawer\_of\_the\_cabinet & [grasp, lift, tilt] \\
87 & KITCHEN\_SCENE6\_close\_the\_microwave & [grasp, move, place, rotate] \\
88 & KITCHEN\_SCENE6\_put\_the\_yellow\_and\_white\_mug\_to\_the\_front\_of\_the\_white\_mug & [grasp, move, place] \\
89 & KITCHEN\_SCENE4\_put\_the\_wine\_bottle\_in\_the\_bottom\_drawer\_of\_the\_cabinet & [grasp, tilt] \\
90 & LIVING\_ROOM\_SCENE6\_put\_the\_chocolate\_pudding\_to\_the\_left\_of\_the\_plate & [grasp, move, place] \\
\bottomrule
\end{tabularx}
\end{table}

\subsection{Training Data Statistics}
\label{appendix:training_data}

\noindent \textbf{Data Collection and Processing.} 
The training dataset is consolidated from three primary sources, each with specific collection and cleaning protocols:
\begin{itemize}
    \item \textbf{LIBERO Datasets:} Simulation data are sourced from the official LIBERO benchmark. We exclusively apply the OpenVLA~\cite{kim2024openvla} data cleaning protocol to these subsets to filter out unsuccessful attempts and redundant segments, ensuring only high-quality successful trajectories are retained.
    \item \textbf{RLBench:} Data are generated using native demonstration scripts, with 300 successful trajectories collected for each of the 10 tasks.
    \item \textbf{Real-world Data:} A total of 160 trajectories were human-collected in a laboratory setting. This includes 30 trajectories for each of the 4 manipulation tasks (Pick sponge, Pick cup, Pick cube in drawer, Pick cube in caddy) and 20 trajectories for each of the 2 simpler \textit{push} tasks.
\end{itemize}

\noindent \textbf{Post-Segmentation Statistics.}
To facilitate primitive-level learning, we apply a rule-based segmentation process to all successful trajectories. Any residual trajectory fragments that fail to meet the primitive termination criteria are discarded to maintain the purity of motion patterns. Table~\ref{tab:detailed_data_stats} summarizes the scale of original trajectories and the final count of valid primitives.

\begin{table}[htbp]
\centering
\caption{Statistics of successful trajectories and resulting primitives across all training datasets.}
\label{tab:detailed_data_stats}
\vspace{1em}
\small
\begin{tabularx}{\textwidth}{X c c c}
\toprule
\textbf{Benchmark} & \textbf{Original Trajectories} & \textbf{Resulting Primitives} & \textbf{Source / Strategy} \\ 
\midrule
Libero-Object      & 454      & 1,368  & Official (OpenVLA Cleaned) \\
Libero-Spatial     & 432      & 1,293  & Official (OpenVLA Cleaned) \\
Libero-Goal        & 428      & 1,121  & Official (OpenVLA Cleaned) \\
Libero-90 Subset   & 3,688    & 10,546 & Official (OpenVLA Cleaned) \\ \midrule
RLBench            & 3,000    & 8,090  & Script-generated (300/task) \\ \midrule
Real-world ID      & 160      & 400    & Human-collected (20-30/task) \\ \midrule
\textbf{Total}     & \textbf{8,162} & \textbf{22,818} & - \\
\bottomrule
\end{tabularx}
\end{table}

\begin{figure}[htbp]
\centering
\begin{tikzpicture}
\begin{axis}[
    ybar,
    enlarge x limits=0.1,
    ylabel={Proportion (\%)},
    symbolic x coords={Grasp, Place, Move, Lift, Push, Press, Insert, Twist, Pull, Tilt, Rotate},
    xtick=data,
    x tick label style={rotate=45, anchor=east},
    nodes near coords,
    nodes near coords style={font=\tiny, /pgf/number format/fixed},
    width=\textwidth,
    height=5.5cm,
    ymin=0, ymax=40,
    bar width=14pt,
    axis x line*=bottom,
    axis y line*=left,
    ymajorgrids=true,
    grid style={dashed,gray!30}
]
\addplot[fill=blue!40, draw=blue!60] coordinates {
    (Grasp,34.2) (Place,23.4) (Move,18.6) (Lift,12.9) (Push,2.2) 
    (Press,2.2) (Insert,2.0) (Twist, 2.0) (Pull, 1.6) (Tilt, 1.4) (Rotate, 0.9)
};
\end{axis}
\end{tikzpicture}
\caption{Average primitive distribution across the consolidated training sets.}
\label{fig:primitive_bar_chart}
\end{figure}
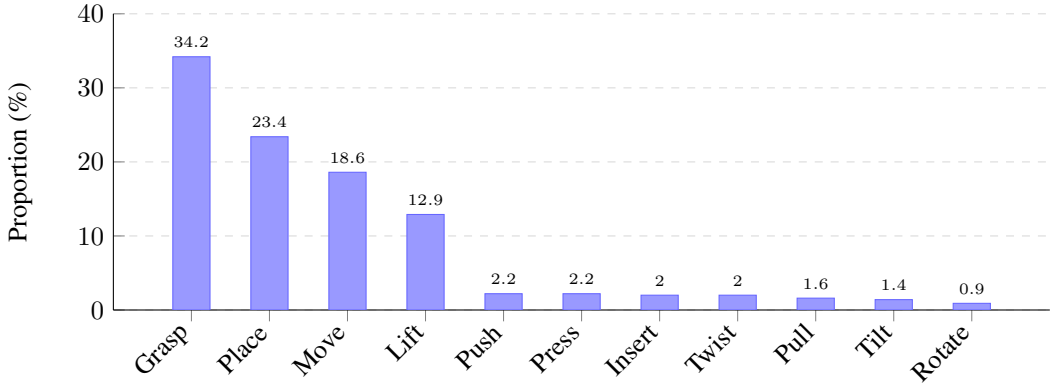

\noindent \textbf{Primitive Distribution.} 
The consolidated training set exhibits a long-tail distribution across 11 primitive categories (Fig.~\ref{fig:primitive_bar_chart}). \textit{Spatial
Transport} primitives (Grasp, Place, Move, Lift) dominate the dataset, providing a robust foundation for basic manipulation, while other more complex primitives (e.g., Tilt, Insert) represent the more challenging sparse distributions in the training data.

\subsection{Model Settings}
\noindent \textbf{Model Configurations.}
We evaluate our framework using three distinct VLA model configurations: \textbf{OpenVLA}~\cite{kim2024openvla}, \textbf{OpenVLA-OFT}~\cite{kim2025optimizing} (Optimized Fine-Tuning variant of OpenVLA), and \textbf{pi0.5}~\cite{intelligence2025pi05}. All models primarily follow their official default configurations, with specific adjustments to batch size, training iterations and action chunk sizes as detailed below:

\begin{itemize}
\item \textbf{OpenVLA:} As a single-step prediction model, we focus on training iterations. For the \texttt{LIBERO-Spatial}, \texttt{Object}, and \texttt{Goal} benchmarks, we follow the official fine-tuning setting of 50k iterations. For \texttt{LIBERO-90} and \texttt{RLBench}, we increase the training to 100k iterations to accommodate the significantly larger dataset scale and increased task complexity. The finetuning command uses a \texttt{batch\_size} of 4 with 4 gradient accumulation steps.
\item \textbf{OpenVLA-OFT (Optimized Fine-Tuning):} This variant employs an optimized fine-tuning strategy tailored for the LIBERO benchmark. Following its specific protocol, we set the training duration to 150k iterations for all simulation tasks (LIBERO and RLBench), with an action chunk size of 5. The instruction parameters are set to a \texttt{batch\_size} of 2 and 4 gradient accumulation steps.
\item \textbf{pi0.5:} Based on the default training scale for LIBERO, pi0.5 is trained for 30k iterations across all simulation and real-world benchmarks. The \texttt{batch\_size} parameter is set to 56. We use an action chunk size of 10 for simulation, which is increased to 20 in real-world experiments to enhance the model's robustness against physical perturbations and environmental noise inherent in live execution.
\end{itemize}

All fine-tuning for OpenVLA and OpenVLA-OFT was conducted on an 8-GPU L40S(48g) server, taking 60 and 100 hours, respectively. Fine-tuning for Pi-0.5 was completed on an 8-GPU A100(80g) server in 20 hours. For LIBERO benchmarks, evaluations were performed on the same hardware used for fine-tuning (L40S for OpenVLA/OpenVLA-OFT and A100 for Pi-0.5). In contrast, all RLBench and real-world experiments were carried out on an RTX 4090.

\FloatBarrier

\section{Additional Discussions}
\label{sec:additional_discussions}

\subsection{Generalization Analysis}
\label{sec:generalization_analysis}

\textbf{PrimitiveVLA enhances zero-shot generalization by transforming complex, out-of-distribution tasks into a composition of familiar, high-confidence primitives.}

\begin{figure}[tbh]
    \centering
    \includegraphics[width=\columnwidth]{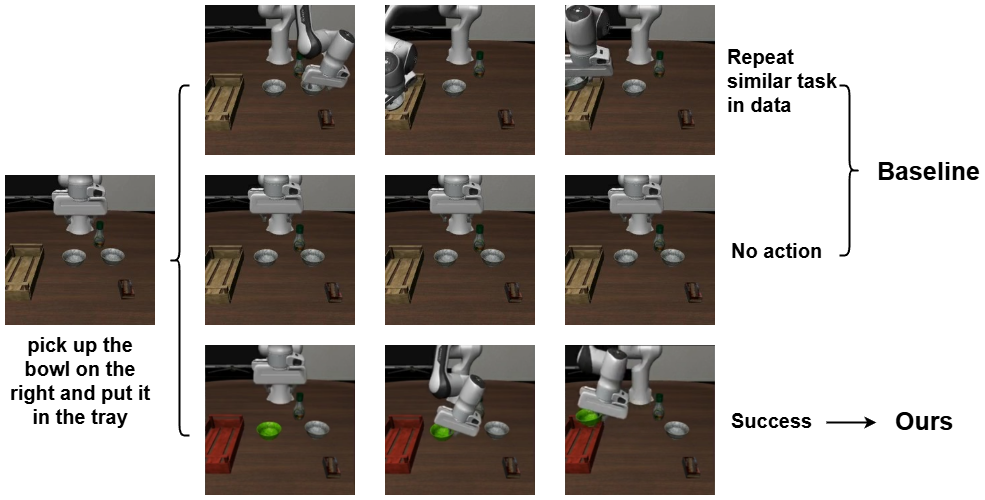}
    \caption{\textbf{Generalization performance examples on an OOD task}: \textit{"pick up the bowl on the right and put it in the tray"}.}
    \label{fig:genelization}
\end{figure}

As shown in Figure \ref{fig:genelization}, standard VLA models often struggle with zero-shot generalization due to two critical failure modes: 
(1) \textbf{Imitative interference}, where the model mistakenly repeats a similar but incorrect task from the training data. For example, when given an OOD task like \textit{"pick up the bowl on the right and put it in the tray"}, a baseline VLA tends to execute a previously learned task, such as \textit{"pick up the bowl on the left and put it in the tray"}, failing to adapt to the new instruction.
(2) \textbf{Task-level ambiguity}, where complex, unseen instructions lead to \textit{"frozen" behaviors} (the robot fails to act entirely); 

PrimitiveVLA effectively addresses these challenges by decomposing novel, complex tasks into a sequence of known, reusable interactions. By applying \textbf{primitive-specific masking} and \textbf{unified primitive instructions}, we map the OOD task to a series of \textbf{primitives} that are already present in the training data—such as \textit{"grasp object with green mask"} or \textit{"place in object with red mask"}. This transformation allows the model to execute completely new tasks simply by composing in-distribution \textbf{motion patterns}.

\definecolor{pieBlueA}{HTML}{AACCFF} 
\definecolor{pieBlueB}{HTML}{DDE5FF} 
\definecolor{pieOrange}{HTML}{FFDDAA} 
\definecolor{pieGrayA}{HTML}{E8E8E8} 
\definecolor{pieGrayB}{HTML}{F2F2F2} 

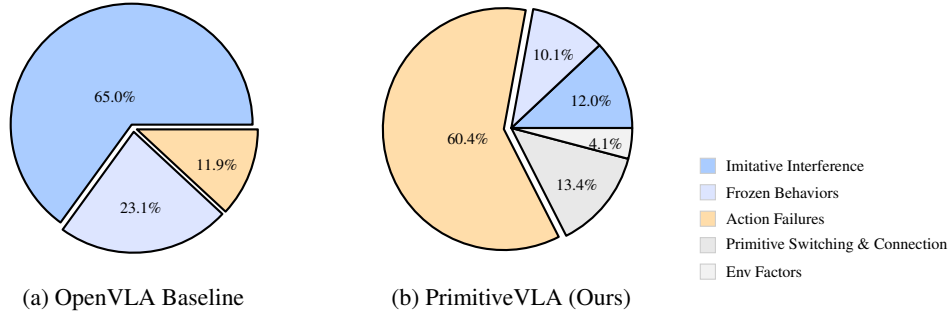
\begin{figure}[tbh]
    \centering
    \begin{tikzpicture}
        
        \begin{scope}
            \pie[
                pos={0,0},
                radius=1.6,
                text=inside,
                color={pieBlueA, pieBlueB, pieOrange},
                explode=0.05,
                font=\tiny,
                after number=\%
            ]{
                65.0/, 
                23.1/,
                11.9/
            }
            \node[below=2.0cm of {0,0}] {\small (a) OpenVLA Baseline};
        \end{scope}

        \begin{scope}[shift={(5.0,0)}] 
            \pie[
                pos={0,0},
                radius=1.6,
                text=inside,
                color={pieBlueA, pieBlueB, pieOrange, pieGrayA, pieGrayB},
                explode={0, 0, 0.1, 0, 0}, 
                font=\tiny,
                after number=\%
            ]{
                12.0/,
                10.1/,
                60.4/,
                13.4/,
                4.1/
            }
            \node[below=2.0cm of {0,0}] {\small (b) PrimitiveVLA (Ours)};
        \end{scope}

        \begin{scope}[shift={(7.5,-0.6)}] 
            \draw[fill=pieBlueA, draw=gray!40] (0,0.0) rectangle (0.2,0.2);
            \node[anchor=west, font=\tiny] at (0.22, 0.1) {Imitative Interference};
            
            \draw[fill=pieBlueB, draw=gray!40] (0,-0.35) rectangle (0.2,-0.15);
            \node[anchor=west, font=\tiny] at (0.22, -0.25) {Frozen Behaviors};
            
            \draw[fill=pieOrange, draw=gray!40] (0,-0.7) rectangle (0.2,-0.5);
            \node[anchor=west, font=\tiny] at (0.22, -0.6) {Action Failures};
            
            \draw[fill=pieGrayA, draw=gray!40] (0,-1.05) rectangle (0.2,-0.85);
            \node[anchor=west, font=\tiny] at (0.22, -0.95) {Primitive Switching \& Connection};
            
            \draw[fill=pieGrayB, draw=gray!40] (0,-1.4) rectangle (0.2,-1.2);
            \node[anchor=west, font=\tiny] at (0.22, -1.3) {Env Factors};
        \end{scope}

    \end{tikzpicture}
    \caption{Failure mode distribution on LIBERO-90-Novel tasks. The results show a structural migration of errors from high-level semantic ambiguity to physical interaction challenges.}
    \label{fig:error_pie}
\end{figure}

\vspace{2pt}
\noindent \textbf{Shift in Error Patterns.} We conducted a granular failure analysis using OpenVLA as the baseline and base model (OpenVLA v.s. OpenVLA + ours) on the LIBERO-90 Novel task suite, shown in Figure \ref{fig:error_pie}.

\begin{itemize}
\item \textbf{Baseline: Dominance of Semantic Failures.} In the baseline configuration, the vast majority of failures are rooted in high-level cognitive confusion. Imitative Interference accounts for 65.0\% of total errors, followed by Frozen Behaviors at 23.1\%. Combined, these "high-level" failures constitute over 88\% of all baseline errors, indicating that the model primarily struggles with what to do when encountering unfamiliar instructions or environments.

\item \textbf{PrimitiveVLA (ours): Transition to Execution-Centric Challenges.} With PrimitiveVLA, the error landscape shifts significantly toward low-level execution. 
\begin{itemize}
    \item \textbf{Significant Reductions:} The proportion of \textbf{Imitative Interference} drops sharply to \textbf{12.0\%}, and \textbf{Frozen Behaviors} decrease to \textbf{10.1\%}. This confirms that task disassembly and primitive-specific masking provide a much clearer goal and execution flow, effectively bypassing task-level ambiguity.
    \item \textbf{Dominance of Action Failures:} Action Failures now become the primary error mode, rising to 60.4\%. This increase is a direct result of the model's higher task-initiation rate;instead of defaulting to similar ID tasks or remaining "frozen," the model now actively attempts OOD tasks. Consequently, the failure profile migrates from high-level cognitive errors to low-level execution challenges. The resulting failures—often seen as trajectory deformations—reflect the challenge of generalizing learned motion patterns to OOD physical distributions (e.g., unseen objects, novel environmental states, or unfamiliar spatial arrangements).
    \item \textbf{Emerging Paradigm-Specific Errors:} New error types unique to our framework, including Primitive Switching (6.5\%) and Motion Connection (6.9\%), account for a combined 13.4\%. These errors highlight the remaining gaps in the heuristic switching logic and the sparse coverage of novel primitive sequences in our current training data.
\end{itemize}
\end{itemize}

\subsection{Data Efficiency Analysis}
\label{sec:data_efficiency}

To investigate why PrimitiveVLA maintains high success rates even under a \textbf{half-scale (50\% data)} regime, we analyze the distribution of 11 identified \textit{primitives} across the Libero benchmarks. The detailed statistics are summarized in Table~\ref{tab:primitive_stats}.

\begin{table}[h]
\centering
\caption{Detailed distribution of primitives across the Libero}
\label{tab:primitive_stats}
\vspace{1em}
\begin{tabular}{lccccc}
\toprule
\textbf{Primitive} & \textbf{Libero-90} & \textbf{Libero-Goal} & \textbf{Libero-Spatial} & \textbf{Libero-Object} & \textbf{Total} \\ \midrule
Grasp              & 86                 & 8             & 10               & 10              & 114            \\
Place              & 63                 & 7             & 10               & 10              & 90             \\
Move               & 49                 & 3             & 10               & 10              & 72             \\
Lift               & 28                 & 4             & 0                & 0               & 32             \\
Insert             & 12                 & 0             & 0                & 0               & 12             \\
Pull               & 5                  & 2             & 0                & 0               & 7              \\
Push               & 8                  & 1             & 0                & 0               & 9              \\
Press              & 0                  & 1             & 0                & 0               & 1              \\
Twist              & 5                  & 1             & 0                & 0               & 6              \\
Tilt               & 6                  & 0             & 0                & 0               & 6              \\
Rotate             & 2                  & 0             & 0                & 0               & 2              \\ \midrule
\textbf{Total Tasks} & \textbf{90}        & \textbf{10}   & \textbf{10}      & \textbf{10}     & \textbf{120}   \\ \bottomrule
\end{tabular}
\end{table}

Even when each benchmark is fine-tuned independently, PrimitiveVLA demonstrates exceptional data efficiency through three core mechanisms:

\begin{itemize}
\item \textbf{PrimitiveVLA achieves high data efficiency by concentrating supervision on a shared set of motion patterns.} In diverse suites like \textit{Libero-90}, traditional VLAs must learn 90 distinct mappings between instructions and trajectories, resulting in extremely sparse data for any specific command. Our framework reshapes this sparse task distribution into a dense action distribution by unifying tasks into a compact set of primitives. For instance, while there are 90 unique task descriptions, the same motion pattern for \textit{grasp} is executed in 86 of them. This allows the model to repeatedly reinforce a few \textbf{high-frequency motion patterns} rather than memorizing 90 isolated tasks, providing much stronger supervision for primitives even with 50\% fewer samples.

\item \textbf{We reduce the learning burden by aligning diverse tasks into a consistent structural format.} In the \textit{Libero-Spatial} and \textit{Libero-Object} benchmarks, the primary challenge lies in environmental variations (e.g., object positions or categories) rather than logical complexity. PrimitiveVLA explicitly simplifies all 20 tasks in these suites into a consistent \texttt{[grasp, move, place]} sequence. This structural alignment shifts the learning burden from understanding multi-stage linguistic logic to mastering short-range primitives. Such simplification enables the model to reach better performance with significantly less data than baselines.

\item \textbf{By treating every segment as an equal training unit, we prevent critical short-duration actions from being ignored.} Standard VLA training typically employs uniform temporal down-sampling on full trajectories. However, some motion patterns like \textit{lift}, \textit{press} often occupy a very short duration within a long trajectory. Uniform sampling tends to drown out these short actions, leading to insufficient learning of these task segments. By isolating these short primitives into \textbf{independent training units}, PrimitiveVLA ensures that every segment gets an equal training weight, making short moves just as important as longer ones. This adaptive temporal resolution prevents the "blurring" of short actions in the long-horizon data, ensuring high success rates even under severe data constraints.
\end{itemize}

\subsection{Robustness to VLM/LLM Noise}
\label{sec:mask_drift_analysis}

\textbf{Errors or hallucinations from the VLM/LLM do not significantly affect our results.} 

In our experiments, we observe that the disassembly process occasionally introduces two types of infrequent noise ($<$5\% in LIBERO, $<$10\% in RLBench): (1) \textbf{primitive merging}, where transitions between adjacent motion phases are blurred, and (2) \textbf{mask drift}, characterized by minor spatial misalignment or incorrect part labeling. Interestingly, our experimental observations indicate that \textit{primitive merging} is typically rectified by the subsequent rule-based logic, while the impact of \textit{mask drift} is effectively mitigated by the VLA’s inherent robustness.

Specifically, although VLM-based masking provides crucial spatial grounding for MCR, the resulting minor spatial variances do not degrade performance in \textbf{In-Distribution (ID)} tasks. Instead, the VLA policy learns to internalize this noise during training, treating slight pixel-level shifts as a form of implicit data augmentation that enhances its overall stability. While the sensitivity to such drift is slightly more pronounced in \textbf{Out-of-Distribution (OOD)} tasks involving novel objects, our experiments confirm that the explicit physical grounding provided by MCR consistently outweighs the noise from minor shifts. This trade-off results in a significant net gain, ensuring robust assembly performance even in the presence of imperfect disassembly or spatial noise.

\subsection{Latency and Computational Overhead}
\label{sec:latency_analysis}
\textbf{The PrimitiveVLA mechanism is designed to be lightweight, ensuring that the inference frequency remains consistent with the base VLA models.}

\noindent The planning of the primitive sequence and the generation of LLM-based switch code are conducted in the Pre-execution planning, which is a common stage for VLA, thus incurring zero overhead during real-time execution. During the rollout, the Python-based switching logic is extremely lightweight.

\noindent \textbf{Tracking Latency.} We utilize Cutie for real-time mask tracking. The latency of Cutie scales with image resolution: approximately \textbf{9ms} per frame at $256 \times 256$ and \textbf{30ms} per frame at $768 \times 768$. To ensure high success rates, we employ $768 \times 768$ resolution to identify and track masks, updating the VLA input once before each action chunk generation.

\begin{table}[htb]
\vspace{-5pt}
\centering
\caption{\textbf{Inference Latency Comparison.} We report the average per-step latency (ms) in Libero, including mask tracking and switching logic.}
\label{tab:latency}
\vspace{1em}
\begin{tabular}{lccc}
\toprule
\textbf{Model} & \textbf{Chunk Size} & \textbf{Baseline Latency} & \textbf{PrimitiveVLA Latency} \\ %
\midrule
OpenVLA        & 1                   & $\sim$500 ms              & $\sim$540 ms                  \\
OpenVLA-OFT    & 5                   & $\sim$88 ms               & $\sim$96 ms                   \\
$\pi_{0.5}$    & 10                  & $\sim$67 ms               & $\sim$72 ms                   \\ 
\bottomrule
\end{tabular}
\end{table}

As shown in Table~\ref{tab:latency}, the additional overhead is negligible. In our current experiments, since the chunk size is relatively small, updating the mask once per chunk is sufficient. For future VLAs with much larger chunk sizes (e.g., $>20$), we suggest adopting \textbf{asynchronous mask updates} or updating every fixed number of steps to maintain the execution frequency.

\end{document}